\documentclass[sigplan, 10pt]{acmart}
\renewcommand\footnotetextcopyrightpermission[1]{}
\AtBeginDocument{%
  }

\usepackage{xspace}
\usepackage[utf8]{inputenc}
\usepackage{comment}
\usepackage{multirow}
\usepackage{amsmath}
\usepackage{color}
\usepackage{tikz}
\usepackage{booktabs}
\usepackage{xcolor, colortbl} 
\usetikzlibrary{shapes, decorations, arrows, calc, arrows.meta,fit, positioning}
\tikzset{
    -Latex, auto, node distance =1 cm and 1 cm, semithick,
    state/.style ={ellipse, draw, minimum width = 0.7 cm},
    point/.style = {circle, draw, inner sep=0.04cm,fill,node contents={}},
    bidirected/.style={Latex-Latex,dashed},
    el/.style = {inner sep=2pt, align=left, sloped}
}

\usepackage{amsfonts}
\usepackage[super]{nth}
\usepackage{xurl}
\usepackage[inline]{enumitem}
\usepackage{amsthm}
\usepackage{mathtools}
\usepackage{thmtools}
\usepackage{thm-restate}
\usepackage{cleveref}
\usepackage{graphicx}
\usepackage{hyperref}
\usepackage{xcolor}
\usepackage{subcaption}
\usepackage{algorithm}
\usepackage{algpseudocode}
\hypersetup{
    colorlinks=true,
    linkcolor=blue,      
    urlcolor=blue,       
    citecolor=green,      
}

\newif\ifdraft
\draftfalse

\newcommand{\name}{{Ascendra}\xspace}

\begin{document}

\title{\name: Dynamic Request Prioritization for Efficient LLM Serving}

\author{Azam Ikram}
\authornote{These authors contributed equally to this work.}
\affiliation{%
  \institution{Purdue University}
  \country{}
}
\author{Xiang Li}
\authornotemark[1]
\affiliation{%
  \institution{Purdue University}
  \country{}
}
\author{Sameh Elnikety}
\affiliation{%
  \institution{Microsoft Research}
  \country{}
}
\author{Saurabh Bagchi}
\affiliation{%
  \institution{Purdue University}
  \country{}
}


\begin{abstract}
The rapid advancement of Large Language Models (LLMs) has driven the need for more efficient serving strategies. In this context, efficiency refers to the proportion of requests that meet their Service Level Objectives (SLOs), particularly for Time To First Token (TTFT) and Time Between Tokens (TBT). However, existing systems often prioritize one metric at the cost of the other.

We present Ascendra, an LLM serving system designed to meet both TTFT and TBT SLOs simultaneously. The core insight behind Ascendra is that a request’s urgency evolves as it approaches its deadline. To leverage this, Ascendra partitions GPU resources into two types of instances: low-priority and high-priority. Low-priority instances maximize throughput by processing requests out of arrival order, but at the risk of request starvation. To address this, Ascendra employs a performance model to predict requests at risk of missing their SLOs and proactively offloads them to high-priority instances. High-priority instances are optimized for low-latency execution and handle urgent requests nearing their deadlines. This partitioned architecture enables Ascendra to effectively balance high throughput and low latency. Extensive evaluation shows that Ascendra improves system throughput by up to 1.7× compared to vLLM and Sarathi-Serve while meeting both TTFT and TBT SLOs.
\end{abstract}

\maketitle
\pagestyle{plain}

\vspace{-4 pt}
\section{Introduction}\label{sec:introduction}
In recent years, large language models (LLMs) have become integral to various applications—from answering user queries to aiding software development. As the demand for LLM-powered services continues to grow, companies are racing to integrate LLM capabilities into their platforms. On the research front, significant effort has been devoted to making LLM serving more efficient. Recent work has focused on improving execution speed~\cite{dao2022flashattention, hong2024flashdecoding++}, optimizing I/O access patterns~\cite{kwon2023efficient}, and ensuring fairness across concurrent requests~\cite{sheng2024fairness}. In this paper, we study the problem of efficient LLM serving and propose a system that achieves high throughput while respecting latency service-level objectives (SLOs), without relying on expensive network interconnects.

\begin{figure}[t]
  \centering
  \begin{subfigure}{\linewidth}
    \centering
    \includegraphics[width=\linewidth]{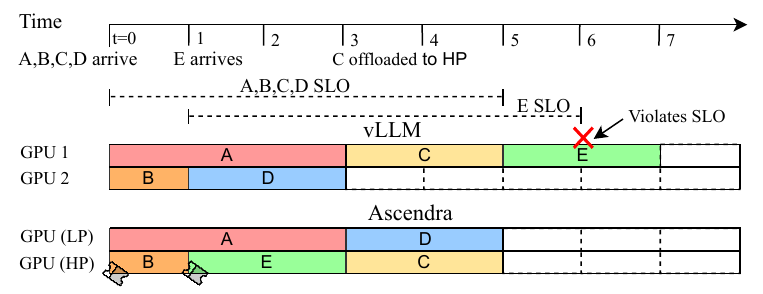}
    \caption{\name offloads urgent requests to high-priority (HP) instances to prevent SLO violations, while baselines like vLLM strictly follow arrival order and result in SLO violation.}
    \label{fig:intro_fig}
  \end{subfigure}
  
  \begin{subfigure}{\linewidth}
    \centering
    \includegraphics[width=\linewidth]{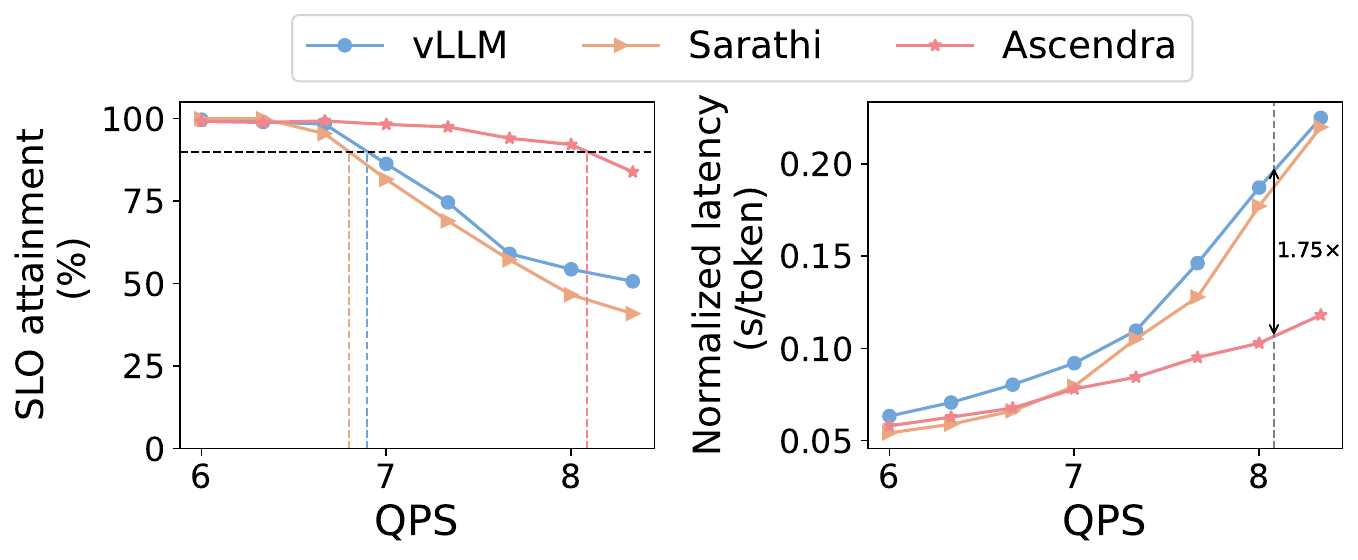}
    \caption{Goodput and normalized throughput when serving Mistral-7B on the \textit{ShareGPT4} dataset using three NVIDIA A100-80G GPUs. \name consistently maintains higher SLO attainment and lower latency compared to existing baselines, even at higher QPS.}
    \label{fig:good_through_put}
  \end{subfigure}
    \vspace{-7mm}
  \caption{\textit{Top}: Out-of-order scheduling and proactive offloading of \name avoid SLO violations. \textit{Bottom}: \name improves SLO attainment and latency compared to baselines.}
  \label{fig:main_figure}
    \vspace{-5mm}
\end{figure}

LLM inference typically consists of two phases: prefill and decode. In the prefill phase, the model processes the input prompt and generates the first token; the latency from request arrival to this token is referred to as Time-to-First-Token (TTFT). In the decode phase, the model autoregressively generates subsequent tokens, with Time-Between-Tokens (TBT) measuring the latency between two consecutive token generations. Real-world applications often impose strict SLOs on both TTFT and TBT. A serving system’s effectiveness is evaluated by its \textit{goodput}—the fraction of requests meeting both SLOs.


Several prior works have explored the trade-off between optimizing TTFT and TBT. For instance, vLLM~\cite{kwon2023efficient} prioritizes prefill to reduce TTFT, at the cost of increased chance of generation stalls. More recently, Sarathi-Serve~\cite{agrawal2024taming} introduced chunked prefill, which improves TBT by interleaving decoding with prefill but suffers from higher TTFT due to frequent I/O access. Another approach, DistServe~\cite{zhong2024distserve}, decouples prefill and decode by running them on separate GPUs. While simple, this disaggregation leads to resource underutilization - prefill is compute-heavy but memory-light, and decode is compute-light but memory-heavy - and requires expensive hardware (e.g., NVLink, NVSwitch, InfiniBand) to transfer large amounts of data (KV caches) between GPUs and nodes.

In this work, we propose a hybrid approach that achieves high goodput without relying on expensive network interconnects. We make three key observations: (1) the delay a request can tolerate before violating its TTFT SLO depends on the prompt length and how long the request has been in the system, (2) First-Come-First-Serve (FCFS) scheduling causes head-of-line blocking, and (3) continuous execution of decoding requests is critical for meeting TBT targets. Based on these insights, we design \name, a system that: (1) assigns dynamically evolving priorities to requests in system based on their proximity to violating the TTFT SLO, (2) enables out-of-order execution to avoid head-of-line blocking, using the current priority as the scheduling policy, and (3) utilizes two distinct types of instances—each optimized for either throughput or latency—to specialize execution based on workload characteristics. This separation enables workload-aware resource allocation and better SLO adherence.

To illustrate the impact of these insights, consider an example involving five requests and a TTFT SLO of five time units as shown in~\autoref{fig:intro_fig}. Under vLLM, the first four requests arrive at time $t = 0$ and are distributed round-robin across two GPUs. Because vLLM processes requests in strict FCFS order, a fifth request (E) arriving at $t = 1$ must wait until all earlier requests have completed, incurring a four-unit delay before starting execution. By the time request E begins its prefill, it has only one time unit remaining to meet its TTFT SLO. However, as the prefill of E is expected to take two time units, E will inevitably violates its TTFT target.

In contrast, consider the same scenario under \name, GPU resources are partitioned into Low Priority (LP) and High Priority (HP) instances. The HP instance accepts a new request only if its queue is empty; otherwise, requests are routed to LP instances. LP instances use an internal value function to determine execution order, optimizing throughput rather than strictly following arrival order. At $t = 1$, request E is forwarded to the idle HP instance. At $t = 3$, the LP instance holds requests C and D, and chooses to execute D while offloading C to HP, as C is projected to miss its TTFT SLO if delayed. This proactive offloading ensures urgent requests are handled promptly, while LP instances maintain high throughput. We validate this behavior in~\autoref{fig:main_figure}, showing that \name improves both throughput and goodput by flexibly managing latency-sensitive requests.

In this paper, we make the following contributions:
\begin{enumerate}
    \item We identify key challenges in existing LLM serving systems, highlighting inefficiencies in handling latency-sensitive workloads in FCFS fashion. We provide an in-depth analysis of the TTFT-TBT trade-off.
    \item We design and implement a hybrid serving architecture that partitions GPU resources into specialized instance types. We introduce a priority-aware scheduling policy that dynamically assigns request priorities at batch-level granularity. This design significantly improves goodput by ensuring timely service for urgent requests while maintaining high system throughput.
    \item We extensively evaluate our system on different models and data sets to show that \name improves goodput compared to state-of-the-art baselines, without requiring additional hardware resources.
\end{enumerate}
\section{Background}\label{sec:background}
\subsection{Phases in LLM Inference}\label{subsec:llm_inference}
LLM serving typically involves two distinct phases: the \emph{prefill phase} and the \emph{decode phase}. Upon receiving a request, the model performs a single forward pass over the entire input prompt, generating a Key-Value (KV) cache and producing the first token — this is the prefill phase. In the subsequent decode phase, the model autoregressively generates one token at a time; each generated token is appended to the input and passed through the model again to produce the next. The decoding phase continues until a stopping condition is met, such as reaching a token limit or encountering an end-of-sequence marker. 

Recent studies~\cite{agrawal2024taming, zhong2024distserve, patel2024splitwise} have shown that these two phases differ in their resource requirements: prefill is compute-bound, while decode is memory-bound. During prefill, the model processes all prompt tokens in parallel, which exposes a high degree of parallelism that can fully saturate GPU compute resources, making it highly compute-intensive. In contrast, the decode phase needs to read the previously generated KV cache and generate one token at a time, requiring frequent memory accesses, which introduces memory bandwidth bottlenecks and limits compute utilization.

\subsection{LLM Scheduling Policies}\label{subsec:scheduling}
Existing LLM scheduling policies can be broadly categorized into three classes based on their primary design goals: \textit{prefill-prioritizing}, \textit{decode-prioritizing}, and \textit{disaggregation-based} strategies.

LLM inference servers such as vLLM~\cite{kwon2023efficient} and Orca~\cite{orca} adopt a prefill-prioritizing strategy that emphasizes responsiveness by ensuring the  generation of the first token, thereby minimizing TTFT. However, frequently interrupting ongoing decode operations to serve new prefill requests can lead to generation stalls~\cite{agrawal2024taming}, potentially degrading throughput.

At the other end of the spectrum, inference frameworks like TensorRT-LLM~\cite{tensorrt_llm} and FasterTransformer~\cite{fastertransformer} employ decode-prioritizing scheduling. These systems prioritize throughput by fully decoding batches of requests before admitting new ones. While efficient for batch execution, this approach can incur long scheduling delays, as new requests must wait for the current requests to complete. To mitigate this, Sarathi-Serve~\cite{agrawal2024taming} introduces a chunked prefill, allowing a partial prefill request to be appended at the end of each decoding batch. While this piggybacking mechanism avoids stalls and maintains token generation continuity, it incurs excessive memory traffic and causes interference between the prefill and decoding phases, extending execution times for both. As this additional latency compounds across all queued requests, it ultimately leads to elevated TTFT, undermining the low TBT guarantee, as requests must satisfy both TTFT and TBT SLOs to count as goodput.

DistServe~\cite{zhong2024distserve} takes an orthogonal approach via disaggregation, decoupling the prefill and decode phases across separate machines. This eliminates intra-node resource contention, enabling both high throughput and low latency without generation stalls. However, this design introduces new challenges: the KV cache generated during the prefill phase — often several gigabytes per request — must be transferred across nodes. To support this level of high-throughput data movement, specialized interconnects such as NVLink and NVSwitch are required. More critically, since machines are now dedicated to a single phase, each becomes a bottleneck along different resource dimensions~\cite{wu2024loongserve} — prefill nodes face increased compute load, while decode nodes contend with high memory pressure, leading to potential under-utilization or new scalability limits.

\subsection{Key-Value Cache Management}\label{subsec:kv_management}
During the decoding phase of LLM inference, key-value pairs computed for prompt and generated tokens must be repeatedly accessed until the completion of the request. To minimize memory access latency, these KV pairs must reside in high-bandwidth memory (HBM) on the GPU. However, even state-of-the-art GPUs such as the A100 and H100 are constrained by an 80 GB HBM capacity. For instance, storing the KV cache for a single token in LLaMA3-8B requires approximately 512 KB—calculated as 2 (key and value) × 4096 (hidden dimension) × 32 (layers) × 2 bytes (FP16 precision). Without efficient memory management, the HBM can be rapidly exhausted by just a few dozen concurrent requests.

To address this, vLLM introduces \textit{PagedAttention}~\cite{kwon2023efficient}, a memory management mechanism that partitions GPU memory into fixed-size blocks. These blocks are dynamically allocated to requests as their decode lengths grow, significantly mitigating memory fragmentation. Llumnix~\cite{sun2024llumnix} further advances this design by supporting live KV migration across inference instances. This approach balances memory usage at the instance level, reduces preemption-induced overheads, and improves overall end-to-end performance.

\subsection{Request Preemption}\label{subsec:request_preemption}
Request preemption occurs when GPU memory is exhausted and decoding cannot proceed without reclaiming resources. To relieve memory pressure and allow other requests to make progress, some ongoing decoding tasks must be temporarily paused and evicted from the GPU HBM. This is typically handled using one of two strategies.

The first approach, used in systems like vLLM and Sarathi-Serve~\cite{kwon2023efficient, agrawal2024taming}, evicts the request by appending its generated tokens to the prompt and placing it back in the waiting queue and release the associated memory. When the request is rescheduled, its KV cache must be recomputed from scratch, incurring recomputation overhead.  The second approach, as adopted by DistServe~\cite{zhong2024distserve}, opts to offload memory instead of recomputation. When memory pressure is high, some ongoing requests are temporarily paused and their KV caches are transferred to host CPU memory. Once sufficient GPU memory becomes available, these requests and their associated KV data are reloaded, and decoding resumes. 

\section{Challenges}\label{sec:motivation}
As LLM requests progress through the two distinct phases of prefill and decode, efficiently scheduling and serving them presents unique system-level challenges.

\subsection{Challenges for Prefill}
The prefill phase consumes significant compute resources, as it requires generating a Key-Value (KV) cache for the entire prompt in a single forward pass. Since prompt lengths vary widely across requests, the execution time and resource demands fluctuate unpredictably. While many systems attempt to minimize TTFT, they often fall short of meeting SLOs due to two core challenges:

\noindent \textbf{Unpredictable prompt length. }
The request prompt length varies widely (\autoref{fig:dataset}), ranging from tens to thousands of tokens, which directly impacts the prefill execution time. Consequently, the available time before a request violates its TTFT SLO is not fixed; it depends on both prompt length and how long the request has been waiting in the system. Despite this variability, many systems rely on an FCFS approach, which can result in some requests completing well before their SLOs are reached, leading to inefficient resource utilization while delaying others.

Instead, we propose scheduling prefill requests \textit{closer} to their TTFT SLOs. Since prompt length serves as a natural predictor of execution time, it can be used to infer request urgency and priority in the prefill phase. By prioritizing longer-running or near-deadline requests, this approach improves overall efficiency and responsiveness. Note that our definition of urgency and priority differs fundamentally from that of Llumnix~\cite{sun2024llumnix}, which classifies the urgency and priority of a request based on application-level SLO. This is a much simpler task to handle as requests from applications with similar SLOs can be grouped and routed to a machine targeting the same objective. In contrast, our approach operates at a finer granularity, leveraging prompt-specific characteristics (i.e., prompt length and the time they spend in the system) rather than coarse application-level SLOs.

\noindent \textbf{Head-of-Line Blocking. }
Since TTFT is primarily dictated by prompt length, a request's prompt length effectively determines its prefill execution time. In an FCFS system, longer requests can block shorter ones that arrive later, a classic case of head-of-line blocking. This inefficiency can lead to substantial delays for short requests, even when sufficient computing resources are available. Addressing this issue requires a scheduling strategy that balances execution order with request urgency and resource constraints.

\begin{figure}[t]
    \centering
    \begin{minipage}{0.49\linewidth}
        \centering
        \includegraphics[width=\linewidth]{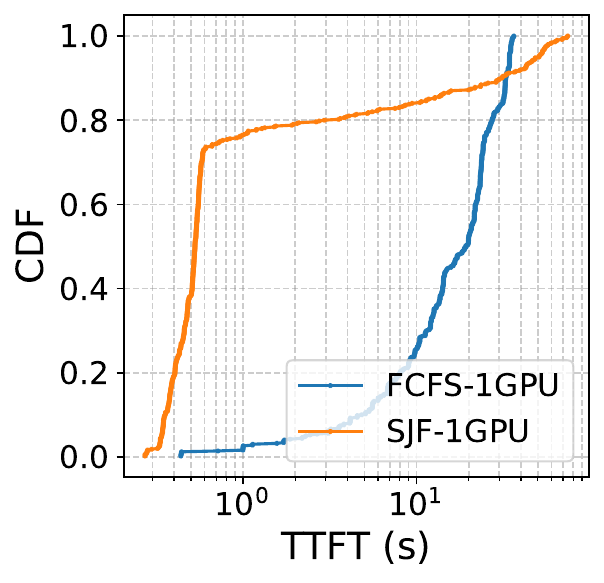}
    \end{minipage}
    \hfill
    \begin{minipage}{0.49\linewidth}
        \centering
        \includegraphics[width=\linewidth]{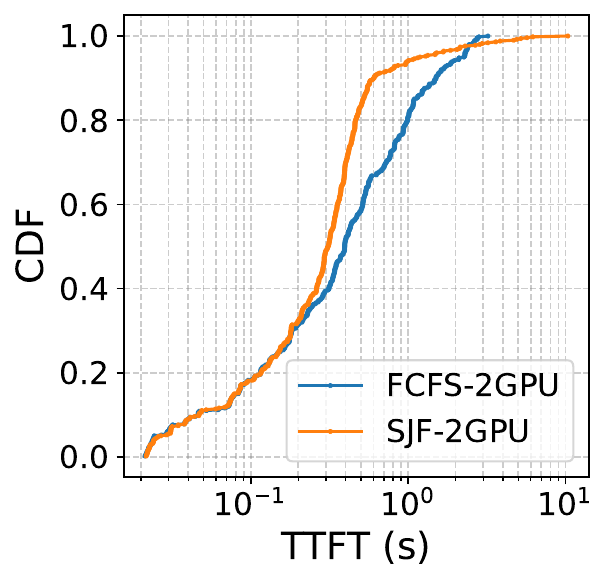} 
    \end{minipage}
    \vspace{-2mm}
    \caption{SJF has lower latency compared to FCFS at the cost of a longer tail. Simply increasing the number of GPUs does not eliminate the tail.}
    \label{fig:fcfs_sjf_ttft}
\end{figure}

\begin{table}[t]
\centering
\resizebox{\columnwidth}{!}{
\begin{tabular}{|l|c|c|c|c|c|}
\hline
 & \multicolumn{3}{c|}{\textbf{Llama-3.1}} & \textbf{Qwen2.5}  & \textbf{DeepSeek-v3} \\ \hline
 \textbf{Model Param \#}                    & 8B       & 70B      & 405B    & 3B       & 37B\footnotemark       \\ \hline
\textbf{KV Cache Size}   & 0.24 GB  & 0.61 GB  & 0.96 GB & 0.068 GB & 3.25 GB    \\ \hline
\end{tabular}
}
\caption{KV cache memory usage for a 2000-token request under FP16 across different models. Larger models consume significantly more memory per request.}
\label{tab:model_kv_cache_usage}
\vspace{-3 mm}
\end{table}
\footnotetext{Number of activated parameters per token.}

However, simply adopting a different scheduling policy does not fully resolve the issue. For example, Figure~\ref{fig:fcfs_sjf_ttft} compares TTFT in vLLM under FCFS and Shortest Job First (SJF). While SJF significantly improves TTFT for most requests, it introduces a pronounced long-tail effect: longer requests are frequently starved, resulting in excessive delays that often violate their TTFT SLOs. Thus, an effective serving system must strike a balance: prioritizing shorter requests to maintain high throughput while ensuring that longer requests are processed in a timely manner, even if they incur slightly higher computational overhead or resource underutilization.

\subsection{Challenges for Decode}
Once the prefill phase completes, the request transitions into the decode phase, where the model autoregressively generates one token at a time until the end of the sequence. We outline the following challenges in the decoding phase.

\noindent \textbf{Unknown Number of Decode Iterations.}
Unlike prefill, where execution time scales with prompt length, decoding is inherently unpredictable: the number of generated tokens varies widely, ranging from tens to thousands (Fig.~\ref{fig:dataset}). This uncertainty makes scheduling difficult: the system cannot determine in advance how long a decoding request will last or how much computational and memory overhead it will introduce. Additionally, decoding workloads must be managed alongside new prefill requests, which may introduce additional demand. Given that decoding typically has stricter SLOs than TTFT, schedulers must dynamically balance prioritization between serving ongoing decode requests and admitting new prefill workloads.

\noindent \textbf{KV-Cache Growth and Memory Pressure.}
While prefill creates a fixed KV-cache matrix from the prompt, decoding appends two vectors (key and value) per token at each step. The cumulative KV-cache can grow rapidly, often reaching several gigabytes for long outputs on large models (Table~\ref{tab:model_kv_cache_usage}), placing sustained pressure on limited GPU memory. This rapid expansion imposes memory pressure, limiting the number of concurrent decoding requests and increasing the risk of memory fragmentation and preemption.


\noindent \textbf{Compute Underutilization. }
Since the KV-cache enables the reuse of previously computed attention values, decoding requires less computation per token than prefill. This shift in workload characteristics allows the scheduler to batch multiple decoding requests together, improving throughput. However, decoding remains highly memory-bandwidth, and memory-space intensive due to frequent KV-cache accesses and updates. As a result, GPU compute tends to be underutilized, while memory bandwidth and capacity become the primary bottleneck as opposed to the prefill phase. As shown in Figure 2, even with just two prefill requests, the GPU’s compute resources are fully saturated, yet memory utilization remains at only 20\%. In contrast, during decoding, the system can batch as many as 128 requests, fully utilizing memory capacity, but at this point, GPU compute usage is around 50\%. This shows that separating prefill and decode onto different GPUs fails to address the core inefficiency — the imbalance between memory and compute utilization.
\begin{figure}[t]
    \centering
    \begin{minipage}{0.49\linewidth}
        \centering
        \includegraphics[width=\linewidth]{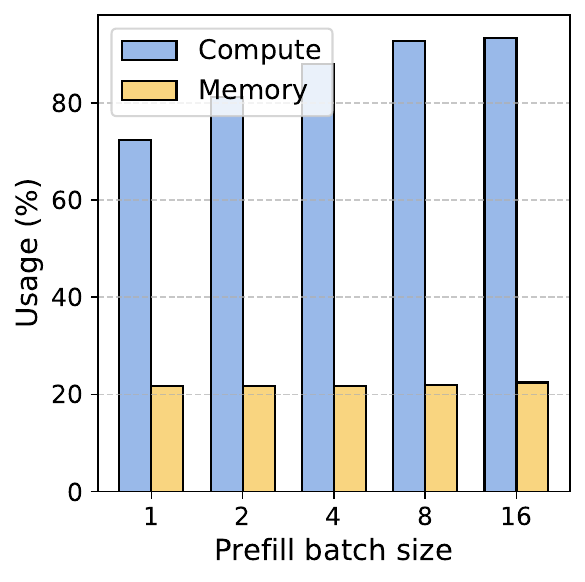}
    \end{minipage}
    \hfill
    \begin{minipage}{0.49\linewidth}
        \centering
        \includegraphics[width=\linewidth]{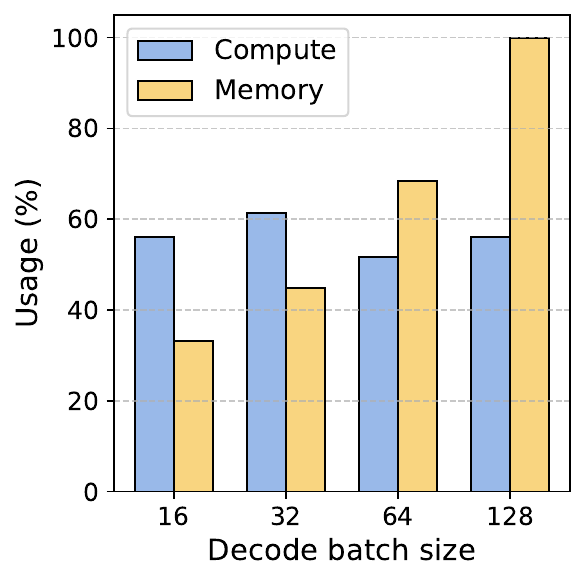} 
    \end{minipage}
    \vspace{-2mm}
    \caption{The prefill phase is compute-bound even with smaller batches, while the decode phase becomes memory-bound, utilizing only ~60\% of compute at a batch size of 128.}
    \label{fig:prefill_decode_profile}
    \vspace{-6 mm}
\end{figure}


\subsection{Insights}
We derive two key insights that guide the design of our serving system, \name, to address the above challenges:

\noindent \textbf{Evolution of Request Priority. } During the prefill phase, the primary objective is to meet the TTFT SLO. However, due to the varying prompt lengths of requests, the amount of time a request can wait before violating its SLO differs. Moreover, as a request waits longer in the system, its remaining slack time shrinks, making it increasingly urgent. To account for this variability, we define a dynamic notion of \textit{priority} or \textit{urgency} for each prefill request. Requests with longer tolerable delays are assigned a lower priority, while those at risk of violating the TTFT SLO within the next few iterations are given higher priority. This priority evolves at batch-level granularity as requests spend more time in the system. We propose that a scheduling system should use this evolving priority to decide which requests should be processed in each batch.

\noindent \textbf{Out-of-Order Prefill Execution. } With the priority of each request defined, we further observe that high-priority requests should be processed before low-priority ones. Since priority depends on both prompt length and time in the system, a longer request may outrank an earlier shorter one. In FCFS systems, this mismatch can delay high-priority requests, causing SLO violations. We propose executing requests out-of-order based on priority, ensuring that urgent requests meet their SLOs even if they arrive later.

Building on these insights, we next discuss the design overview of \name. The goal of \name is to maximize the throughput of the LLM serving system while ensuring that requests meet their SLOs.

\section{Design Overview}
Based on our observations, no single scheduling policy performs optimally across all scenarios. To address this, we propose a novel scheduling framework that partitions the serving infrastructure into two components, each governed by a distinct scheduling policy tailored to a specific objective-minimizing latency for interactive requests (TTFT) and maximizing throughput. We introduce the abstraction of an \textit{instance}, defined as a cluster of GPUs capable of jointly processing a single request. An instance may contain as few as one GPU if the model fits within its memory. With model parallelism such as tensor parallel (TP) or pipeline parallel (PP), however, an instance can span multiple GPUs. To optimize performance, we classify instances into two categories: low-priority (LP) and high-priority (HP), forming distinct instance pools. This separation allows the scheduler to make finer-grained decisions, such as reserving fast paths for latency-sensitive requests while still maintaining high throughput for background workloads. This two-tiered instance abstraction forms the foundation for our adaptive scheduling mechanism.

\autoref{fig:overview} illustrates \name, which comprises a controller and two instance pools: HP and LP. The controller serves as the system entry point, initially routing all user API requests to LP instances. Within each pool, requests are assigned to instances in a round-robin manner. LP instances are optimized for throughput but may struggle to meet SLOs under heavy load. If an LP instance detects an imminent SLO violation, it alerts the controller, which promptly offloads the request to an HP instance. HP instances prioritize adherence to deadlines, reducing the likelihood of SLO violations at the cost of slightly lower throughput.

\begin{figure}[t]
  \centering
  \includegraphics[width=\linewidth]{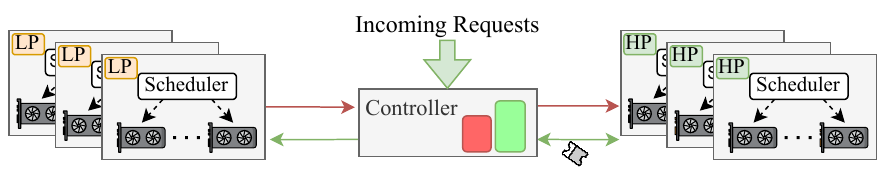}
  \caption{Overview of \name. Green arrows indicate the flow of non-urgent requests, while red arrows represent offloading from LP to HP instances. The controller routes incoming requests to LP instances by default, and to the HP instance when it is available or when a request is urgent.}
  \label{fig:overview}
\end{figure}

Offloading occurs exclusively during the prefill phase, enabling seamless migration by transferring only the prompt and avoiding KV-cache movement. This minimizes data transfer overhead and eliminates the need for specialized high-bandwidth hardware such as NVLink. To meet both TTFT and TBT SLOs, LP instances must balance admitting new requests with ensuring ongoing decoding completes within the TBT deadline. However, when overloaded, LP instances may delay prefill completion due to heavy decoding workloads. To address this, LP instances offload requests to HP instances, which prioritize prefill over decoding. This strategy helps meet TTFT deadlines, though it may occasionally cause minor decoding stalls on HP instances.

We draw inspiration from a related problem in workload allocation between private clusters and public clouds~\cite{luo2024starburst, wu2024can}. Private clusters offer cost efficiency but limited capacity, while public clouds provide scalable compute to meet strict deadlines at a higher cost. Similarly, in our setting, LP instances focus on maximizing throughput while meeting SLOs under normal load. However, under high load, some requests risk starvation. In such cases, the system offloads these requests to HP instances, which prioritize meeting TTFT SLOs-even at the cost of slightly lower throughput.

By combining LP and HP instances, our framework strikes a balance between throughput and latency. This separation of instance types allows us to define specialized goals for each, collectively outperforming existing scheduling schemes. Additionally, operators can dynamically adjust the ratio of HP to LP instances based on workload characteristics, whether to prioritize stringent TTFT or TBT SLOs or adopt a balanced approach. This adaptability allows \name to maintain robust performance across a wide range of system loads. Next, we discuss the design of LP and HP instances in more detail.
 
\section{Low Priority Instance}\label{sec:lp_instance}

In this section, we describe the design of the LP (Low Priority) instance in detail. An effective serving system must aim to meet the SLOs of incoming requests. In our context, each request is subject to two SLO components: TTFT and TBT. Existing serving systems either provide no guarantees on SLO adherence (e.g., vLLM~\cite{kwon2023efficient} or Sarathi~\cite{agrawal2024taming}, which primarily optimize for either TTFT or TBT), or they enforce both SLOs at the cost of significant GPU underutilization (DistServe~\cite{zhong2024distserve}) and specialized interconnects. To address this, we design our LP instance as a high-throughput serving tier. The LP instance prioritizes maximizing the number of generated tokens per unit of time (i.e., throughput) while ensuring that each request meets both TTFT and TBT SLOs.

\begin{figure}[t]
  \centering
  \includegraphics[width=\linewidth]{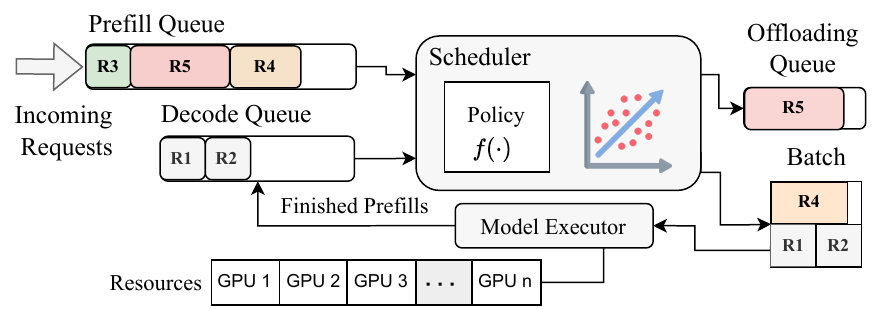}
  \caption{Detailed design of an LP instance. The system maintains separate queues for waiting prefill and ongoing decode requests. At each step, the scheduler uses a policy function and performance model to select requests from both queues and form a batch. It also flags requests at risk of SLO violation, placing them in an offloading queue for the controller to transfer to an HP instance. After execution, completed prefill requests are moved to the decode queue.}
  \label{fig:overview}
\end{figure}

\subsection{Performance Model}
A key component of an LP instance is the ability to accurately estimate the execution time of prefill and a single decoding iteration of the LLM model for a given request. This estimate is essential for guiding scheduling decisions. However, the actual runtime of a request within a batch may deviate from its standalone estimate due to interference with other jobs in the batch, which may include both prefill and decode stages.

Since the execution time of an LLM forward pass is relatively predictable~\cite{zhong2024distserve, narayanan2023cheaply, zhu2024nanoflow}, we construct a performance model to predict the runtime of both the prefill time of individual requests and complete batches. We begin by analytically modeling the computational operations (FLOPs) and memory accesses required to process a batch. These quantities can be derived directly from the model architecture. Given an LLM model (i.e LlaMA 3) with hidden size \textbf{h}, number of heads \textbf{n}, head size \textbf{s}, feed forward intermediate size \textbf{m} and blocksize \textbf{b} (we used flashattention~\cite{dao2022flashattention} as backend for attention computation), we perform the following analysis. For simplicity, we only present the analysis of the attention calculation and skipped the general matrix multiply (GEMM) operations and denote the corresponding flops and memory access-related GEMM operations as $f(GEMM), M(GEMM)$. Further details of this analytical modeling are provided in Appendix~\ref{app:analytical_modeling}.

Suppose that a batch consists of $B_{p}$ prefill requests and $B_{d}$ decode requests. We define $l_{i}$ as the number of prompt tokens for request $i$. The memory access (M) and computation flops (F) for the requests for one attention head in the batch can be expressed as 
\begin{equation}
    M_{p}=\sum_{i=0}^{B_{p-1}} 2l_i s + 3l_i s \left( \frac{l_i}{b} \right), F_{p}=\sum_{i=0}^{B_{p-1}} 2 s l_i^{2}
\end{equation}
Similarly, for the $B_{d}$ decode requests in the batch, we define $\hat{l}_{i}$ as the sum of prompt token and generated token up to the currency batch, the corresponding memory access and computation flops can be represented as
\begin{equation}
    M_{d}=\sum_{i=0}^{B_{d-1}} 2\hat{l}_i s + 2 s, F_{d}=\sum_{i=0}^{B_{d-1}} 2 \hat{l}_i s
\end{equation}
Therefore, to perform an inference on a hybrid batch, the total memory access and computation flops needed  for one decoding layer can be represented as
\begin{align}
    M = (M_{p} + M_{d})\cdot n + M(GEMM) \nonumber \\
    F = (F_{p} + F_{d})\cdot n + F(GEMM)
\end{align}
The time needed for memory access $t_{M}$ and computation $t_{F}$ can be derived through the hardware computation capacity $F_{H}$ and the memory bandwidth capacity $M_{H}$ (that is, 312 TFLOPS and 2 TB/s for NVIDIA A100-80G). In practice, computation can begin only after the relevant data is transferred from high-bandwidth memory (HBM) to SRAM. Depending on the workload and data dependencies, this data transfer may partially or fully overlap with computation. Therefore, we use the sum $(t_{M} + t_{F})$ (representing no overlap) and maximum $max(t_{M}, t_{F})$ (representing full overlap) to capture the two ends. Along with the individual terms $t_{M}=\frac{M}{M_{H}} , t_{F}=\frac{F}{F_{H}}$ to perform a linear regression to find the corresponding parameters $C_{1}, C_{2}$, $C_{3}$, $C_{4}$ and $C_{5}$ to predict batch latency across a range of model configurations and batch compositions.
\begin{align}
    t = C_1 \cdot (\frac{M}{M_{H}} + \frac{F}{F_{H}}) + C_2 \cdot max(\frac{M}{M_{H}},\frac{F}{F_{H}}) \\
    +\ C_3 \cdot\frac{M}{M_{H}} + C_4\cdot\frac{F}{F_{H}} + C_5
\end{align}

\begin{figure}[t]
    \centering
    \begin{minipage}{0.49\linewidth}
        \centering
        \includegraphics[width=\linewidth]{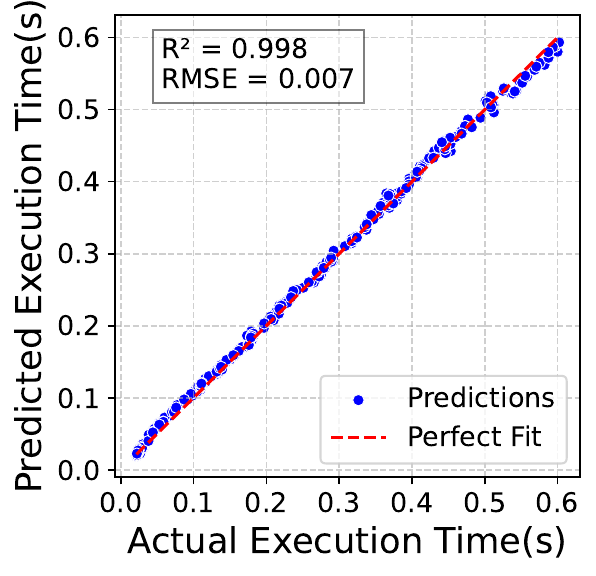}
    \end{minipage}
    \hfill
    \begin{minipage}{0.49\linewidth}
        \centering
        \includegraphics[width=\linewidth]{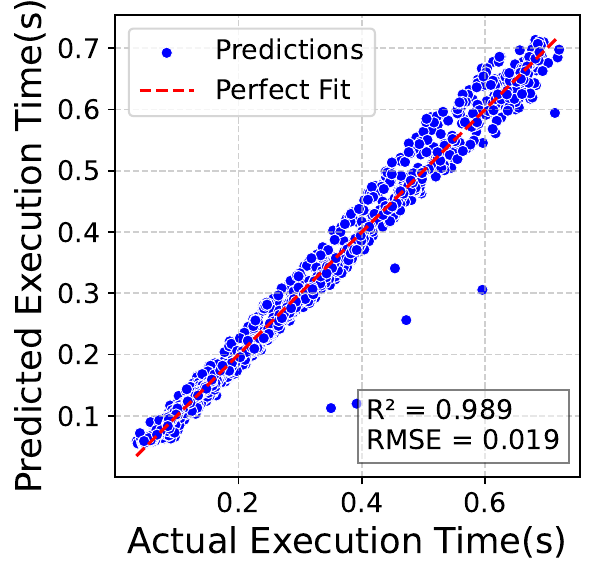} 
    \end{minipage}
    \vspace{-3mm}
    \caption{Accuracy of regression model for predicting execution time of prefill batches (left) and hybrid batches (right). Prediction error remains below 10\% in most cases.}
    \label{fig:batch_regression_model}
    \vspace{-6mm}
\end{figure}

To refine this model in practice, \name profiles the actual execution time of requests at runtime. For each batch, the system logs the number of prefill and decode tokens, along with the observed execution latency. This information is then used to incrementally train and update the performance model. As shown in \autoref{fig:batch_regression_model}, \name requires only five minutes of data collection to build a performance model with less than 10\% error. This data collection process is integrated into the normal serving workflow: the system continuously gathers data from incoming requests, and periodically updates the performance model.

\subsection{Out-of-Order Execution}\label{subsec:out_of_order}

Building on the earlier discussion of scheduling inefficiencies, most existing serving systems~\cite{kwon2023efficient,agrawal2024taming,zhong2024distserve} process incoming requests in a First-Come, First-Served (FCFS) fashion. However, strict FCFS scheduling does not always align with the system's throughput or SLO objectives. For example, if the goal is to defer prefill computation until closer to its associated deadline (prefill SLO), then FCFS may schedule the request too early, resulting in inefficient GPU utilization.

This motivates the adoption of out-of-order scheduling in the LP instance. While FCFS is traditionally used to ensure fairness, we argue that fairness becomes less relevant when every request is associated with a deadline. In fact, strict ordering may hinder overall throughput by prioritizing low-urgency requests over more critical ones.

To enable smarter scheduling decisions, \name prioritizes requests using a \textit{value function} rather than their arrival time [Algorithm.\autoref{alg: value_func}]. This prioritization supports dynamic scheduling policies that better align with system goals. \name is designed to be flexible, supporting a range of policies including Shortest Job First (SJF), FCFS, and Longest Job First (LJF), or any customized policy depending on the operator's preferences. In our implementation, we adopt Earliest Deadline First (EDF) as the default scheduling policy. EDF prioritizes requests nearing their SLO deadline, reducing the likelihood of violations and improving overall system throughput. While our evaluation focuses on EDF, \name supports pluggable scheduling policies to accommodate diverse workloads. We offer an ablation study of two different scheduling policies in \autoref{subsec:order_policy}.

\begin{algorithm}[t]
\caption{Out Of Order Scheduling}\label{alg:outoforder}
\begin{algorithmic}[1]
    \Require waiting queue $W$, compute budget $C$, memory budget $M$, token budget $N$, value policy $\pi$, regression model $\Phi$
    \Ensure List of selected requests

    \State \textbf{for} each request $w_i$ in $W$ \textbf{do}
        \State \hspace{1em} $w_i.\text{val}, w_i.C, w_i.M \gets \pi(\Phi, w_i), \Phi(w_i), \Phi(w_i)$

    \State Sort $W$ by $w_i.\text{val}$ in descending order
    \State $selected\_items \gets []$

    \State \textbf{for} each request $w_i$ in $W$ \textbf{do}
        \State \hspace{1em} \textbf{if} $C > 0$, $M > 0$, $N > 0$ \textbf{then}
            \State \hspace{2em} \textbf{if} $C > w_i.C$, $M > w_i.M$, $N > w_i.prefill$ \textbf{then}
                \State \hspace{3em} Append $w_i$ to $selected\_items$
                \State \hspace{3em} $C = C - w_i.C$
                \State \hspace{3em} $M = M - w_i.M$
                \State \hspace{3em} $N = N - w_i.prefill$
            \State \hspace{2em} \textbf{else}
                \State \hspace{3em} \textbf{break}

    \State \Return $selected\_items$
\end{algorithmic}
\end{algorithm}

\subsection{Request Offloading}

With the out-of-order scheduling scheme, \name selects which requests to execute based on their priority. However, this approach may lead to request starvation, where certain requests are continuously deferred due to incoming more urgent requests. To prevent this, \name includes a request offloading mechanism that transfers urgent, unscheduled requests to HP instances. Any request that has not yet started prefill execution is eligible for offloading.

To make offloading decisions proactively, we account for the worst-latency case scenario in which an HP instance handles a batch with the maximum number of tokens. This helps estimate whether a request can still meet its SLO after being offloaded. The system uses a tunable threshold to decide when a request is too close to its SLO to stay in the waiting queue of LP. If the offload threshold is set too low, the LP may offload requests too late, leaving them little time to finish in HP and increasing the chance of SLO violations. If the threshold is too high, LP may offload requests that could have been handled on time, unnecessarily overloading HP instances and hurting overall throughput. The balance between these two is critical to overall system performance.

\subsection{Further Design Considerations}
To maximize throughput in the LP instance, we adopt a hybrid batching strategy that processes decodes of some requests alongside the prefill of others, a technique known as \textit{piggybacking decodes}~\cite{agrawal2024taming}. In our design, once a request completes its prefill phase, it is added to the set of active decoding requests. At each scheduling iteration, the scheduler batches as many pending decoding requests as possible, subject to the batch size constraint. If there is remaining space in the batch after adding all decoding requests, we utilize our value function to select and pack prefill requests into the batch. This approach ensures continuous progress on decoding, prevents generation stalls, and simultaneously enables the admission of new prefill requests.

\section{High Priority Instance}~\label{sec:hp_instance}
In this section, we discuss the design of HP instance. To ensure the timely execution of urgent requests, HP instances are provisioned to handle prefill workloads offloaded from LP instances. These requests are typically close to their TTFT SLOs and require immediate processing. Upon arrival, offloaded requests are placed in a minimal-length waiting queue to enable near-instant execution.
\begin{figure}[t]
  \centering
  \includegraphics[width=\linewidth]{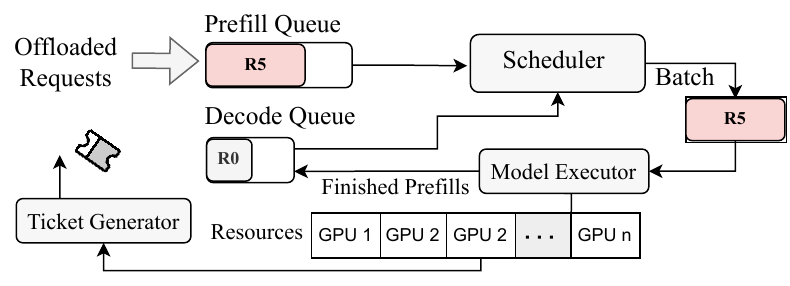}
  \caption{Detailed design of an HP instance. The system admits requests by issuing a ticket to the controller when no prefill is pending, or by accepting urgent offloaded requests. At each step, the scheduler prioritizes prefill over decode to meet TTFT SLO. Once a request completes its prefill phase, it is moved to the decode queue for further processing.}
  \label{fig:overview}
\end{figure}

\subsection{Ticket Entry}
Due to its on-demand nature, HP instance may remain underutilized, especially under light workloads when LP instances can meet all SLOs. To improve utilization. To improve utilization and preserve on-demand responsiveness, we introduce a ticket-entry mechanism. When no requests are offloaded to the HP instance (i.e., the waiting queue is empty), HP instance issues a "ticket" to the controller, prompting it to route the next incoming request directly to HP instance. This opportunistic request helps keep the GPU active. To preserve HP availability for urgent offloads, only one ticketed request is allowed at a time. If an actual offload arrives soon after, it experiences minimal queuing delay — at most one batch cycle. This ticket-entry mechanism can improve the HP utilization under light workload by up to  50\%.

\subsection{Elastic Batch Size}
Existing LLM serving systems typically use a fixed configurable batch size, leading to suboptimal performance due to varying request characteristics. A small batch size may delay prefill requests unnecessarily, despite available memory and compute resources, increasing queuing latency and reducing GPU utilization. However, it improves resilience against growing decoding tokens, which consume significant GPU memory. Conversely, a large batch size may accommodate more prefill requests under high system load, but risks rapid GPU memory consumption during the following decoding phase, leading to frequent preemption and degrade overall system throughput. To address this, an optimal batching strategy must dynamically balance request admission. We achieve this by maintaining a low batch size (i.e., 128) on LP instances to preserve memory for decoding with minimal preemption. On HP instances, we adopt an elastic batch size, allowing prefill to continue as long as GPU memory permits. This is achieved by continuously monitoring the decoding length history and using it to adjust the GPU memory reserved for decoding dynamically. Although this allows more offloaded requests to achieve their TTFT SLO, it risks frequent preemption when the system is under high pressure. 

\subsection{Serving under Load Fluctuation}

When load fluctuations temporarily exceed system capacity, FCFS results in a large number of SLO violations. This occurs because the system spends time processing backlogged requests that have already exceeded their SLO, causing subsequent requests to also miss their deadlines. One fundamental solution is horizontal scaling by adding more instances. However, dynamically scaling GPU nodes is challenging, particularly with specialized hardware requirements (i.e. NVLink, NVSwitch) that must be considered for certain frameworks (i.e. DistServe~\cite{zhong2024distserve}). We propose selectively dropping requests when load fluctuations temporarily exceed system capacity. Such a solution has been adopted in prior DNN serving systems such as Shepherd~\cite{shepherd}. By allowing dropping requests, the system can consistently maintain its goodput when the load temporarily surpasses system capacity.
\section{Implementation}
We implemented \name on top of the open-source LLM serving system vLLM~\cite{kwon2023efficient} and Sarathi-Serve~\cite{agrawal2024taming}. FlashAttention~\cite{dao2022flashattention} and FlashInfer~\cite{flashinfer} kernels are used as the attention computation backend. Our implementation introduces a centralized controller that supports customizable scheduling policies and coordinates request routing across LP and HP instances. At the instance level, we extend the scheduler to support out-of-order execution via pluggable value functions. Similar to Sarathi-Serve~\cite{agrawal2024taming}, NCCL~\cite{NVIDIA_NCCL} is used for both tensor and pipeline parallel communication. The overall system, including HP and LP instance, inter-instance communication, scheduling layer, compute and memory estimation module, and orchestration layer is implemented with 4.1K lines of Python code. Additionally, we develop a batch-level simulation framework with 2.8K lines of Python to enable fine-grained optimal configuration search. 

\section{Evaluation}\label{sec:evaluation}
\subsection{Applications and Dataset}
We evaluate \name on a diverse set of popular models and datasets that reflect a range of real-world serving scenarios. As for baselines, we compare against vLLM and Sarathi-Serve (referred to as Sarathi), which are designed to optimize either TTFT or TBT individually.
We do not directly compare with DistServe~\cite{zhong2024distserve} and Llumnix~\cite{sun2024llumnix} due to their reliance on additional high bandwidth transmission hardware for KV cache transfer (i.e. NVLink or NVSwitch). In this section, we discuss our main results but we also provide further results in Appendix~\ref{app: appendix_result}

\begin{figure}[t]
    \centering
    \includegraphics[width=.98\columnwidth]{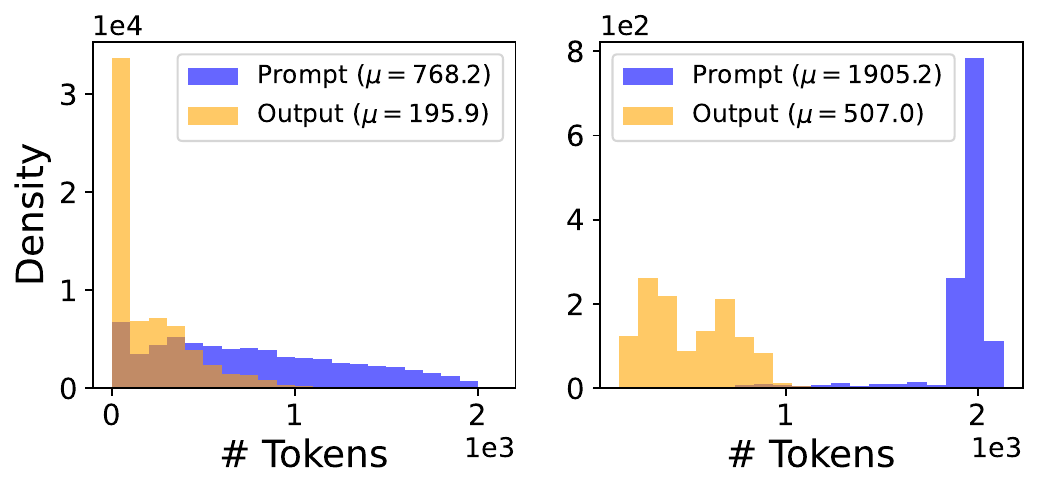}
    \caption{Prompt and response length distribution of \textit{Openchat-ShareGPT4} (left), and \textit{LongBench} (right) dataset.}
    \label{fig:dataset}
\end{figure}

To simulate realistic LLM workloads, we use two publicly available datasets (\textit{Openchat-ShareGPT4} and \textit{Longbench}) that characterize different prompt and output token lengths. We show the distribution of prompt length and output length for each dataset in \autoref{fig:dataset}.

\begin{table}[t]
    \centering
    \renewcommand{\arraystretch}{1.3}
    \setlength{\tabcolsep}{6pt} 
    \resizebox{\columnwidth}{!}{%
        \begin{tabular}{ccccc}  
            \hline
            \textbf{Model} & \textbf{Configurations} & \textbf{TTFT (sec)} & \textbf{TBT (sec)} & \textbf{Dataset} \\ 
            \hline
            \begin{tabular}{c}
                Mistral-7B \\
                Llama3.1-8B
            \end{tabular} 
            & 3 A100-80G
            & \multicolumn{1}{c|}{ 
                \begin{tabular}{c} 
                    1 \\ 2.5 
                \end{tabular} 
            }
            & \begin{tabular}{c} 
                0.15 \\ 0.15
            \end{tabular} 
            & \begin{tabular}{c} 
                ShareGPT~\cite{sharegpt} \\ LongBench~\cite{bai-etal-2024-longbench} 
            \end{tabular} \\ 
            \midrule  
            Qwen-14B
            & 3 A100-80G
             & \multicolumn{1}{c|}{ 
                \begin{tabular}{c} 
                    1.5 \\ 3.0 
                \end{tabular} 
            }
            & \begin{tabular}{c} 
                0.15 \\ 0.15
            \end{tabular} 
            & \begin{tabular}{c}
                ShareGPT~\cite{sharegpt} \\ LongBench~\cite{bai-etal-2024-longbench}
            \end{tabular} \\
            \hline
        \end{tabular}
    }
    \caption{Models, datasets, and corresponding latency SLO in evaluation alongside the testbed setup.}
    \label{tab:model_comparison}
\end{table}

\noindent \textbf{Openchat-ShareGPT4~\cite{sharegpt}} contains the chat conversations from ChatGPT-4~\cite{ChatGPT}. Each interaction with the chatbot is featured as an individual request. For responsiveness, we set the TTFT SLO to be 1 second for LLama3.1-8B and Mistral-7B, and 1.5 seconds for Qwen-14B. The TBT SLO is set at 0.15 seconds, corresponding to a throughput of approximately 400 words per minute—roughly $2-3\times$ faster than the average human reading speed~\cite{BRYSBAERT2019104047}. 

\noindent \textbf{LongBench~\cite{bai-etal-2024-longbench}} contains summarization tasks. It showcases an instance of using LLMs to generate the summary of long documents. It features long prompts with relatively low-variance output lengths. Since responsiveness is less critical in this context, we relax the TTFT SLO to 2.5 seconds for LLama3.1-8B and Mistral-7B, and 3.0 seconds for Qwen-14B, while continuing to enforce a stringent TBT SLO to evaluate throughput efficiency.

To simulate a realistic workload arrival pattern, we generate request arrivals using a Poisson distribution with a varying arrival rate.

\noindent \textbf{Environment Setup:}
We evaluate three popular models Mistral-7B, LLaMA3.1-8B, and Qwen-14B. We use a testbed of 3 80G NVIDIA-A100 GPUs
The nodes are interconnected with a 100 Gbps Infiniband/Ethernet connection.

\subsection{Goodput Comparison}
\begin{figure*}[t]
    \centering
    \begin{minipage}{0.32\textwidth}
        \centering
        \includegraphics[width=\linewidth]{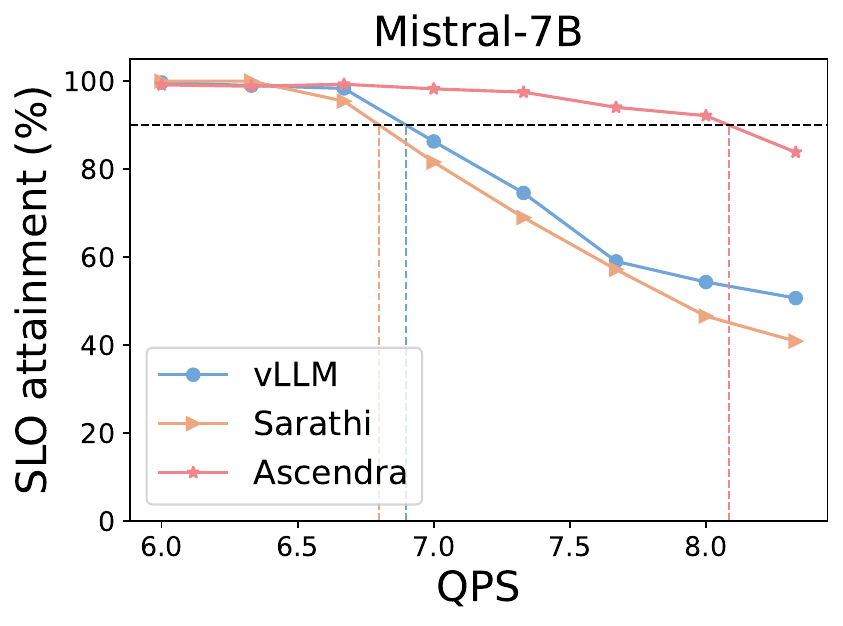}
        \label{fig:mistral_gpt}
    \end{minipage}
    \hfill
    \begin{minipage}{0.32\textwidth}
        \centering
        \includegraphics[width=\linewidth]{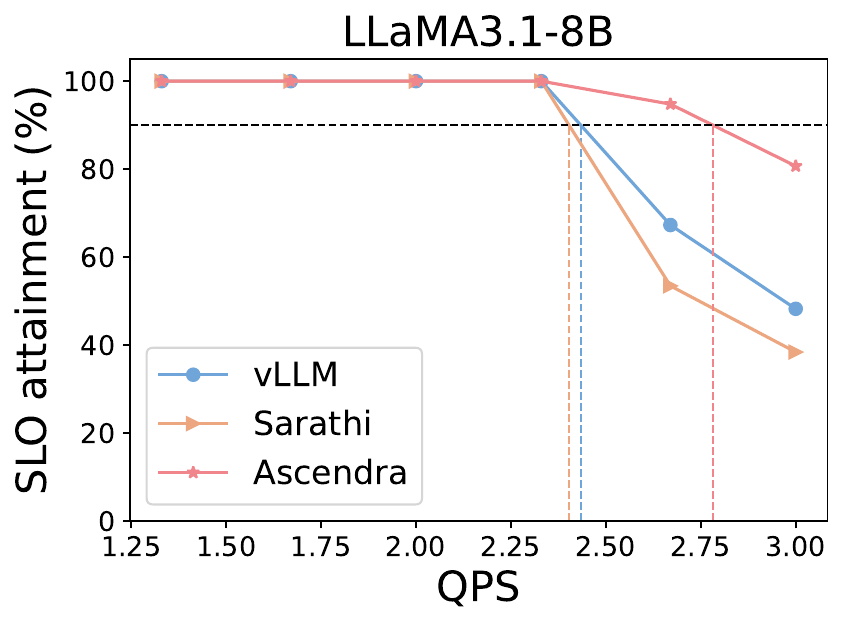}
        \label{fig:llama_longbench}
    \end{minipage}
    \hfill
    \begin{minipage}{0.32\textwidth}
        \centering
        \includegraphics[width=\linewidth]{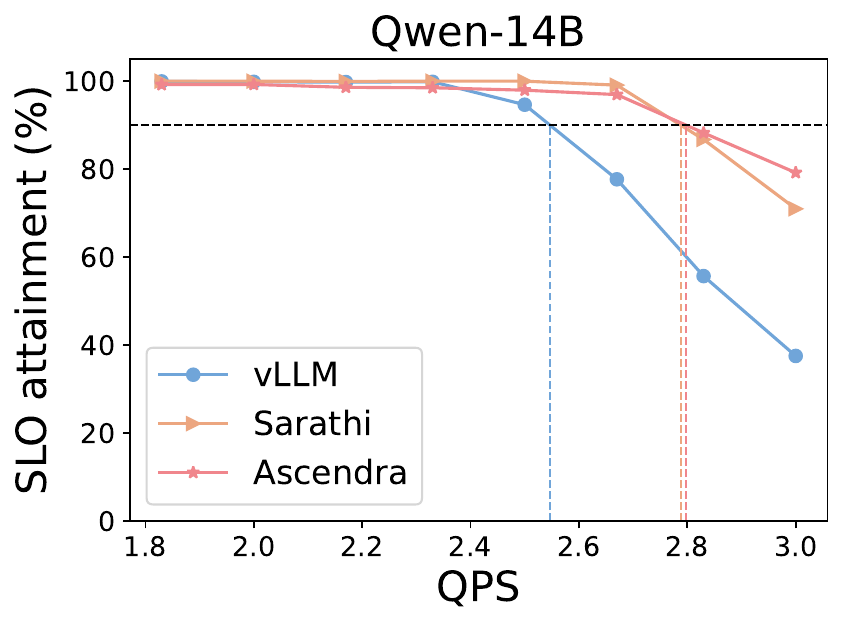}
        \label{fig:qwen_gpt}
    \end{minipage}
    \begin{minipage}{0.32\textwidth}
        \centering
        \includegraphics[width=\linewidth]{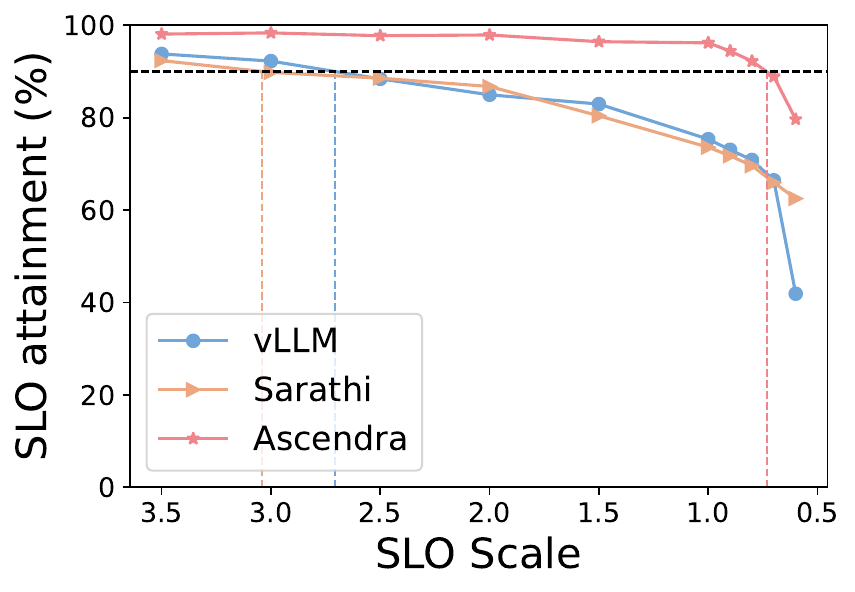}
        
        {\small Mistral-7B (\textit{Openchat-ShareGPT4})}
        \label{fig:mistral_gpt_scale}
    \end{minipage}
    \hfill
    \begin{minipage}{0.32\textwidth}
        \centering
        \includegraphics[width=\linewidth]{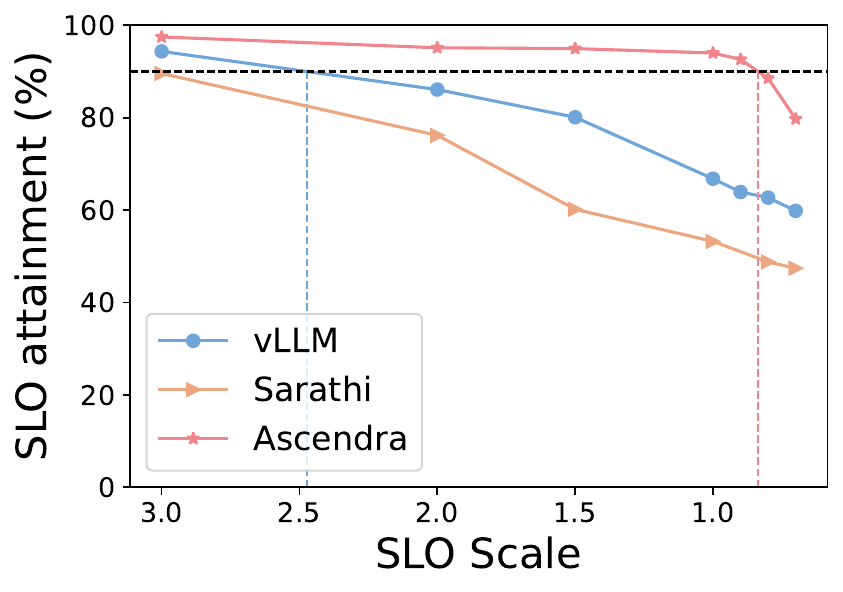}
        
        {\small LLaMA3.1-8B (\textit{LongBench})}
        \label{fig:llama_longbench_scale}
    \end{minipage}
    \hfill
    \begin{minipage}{0.32\textwidth}
        \centering
        \includegraphics[width=\linewidth]{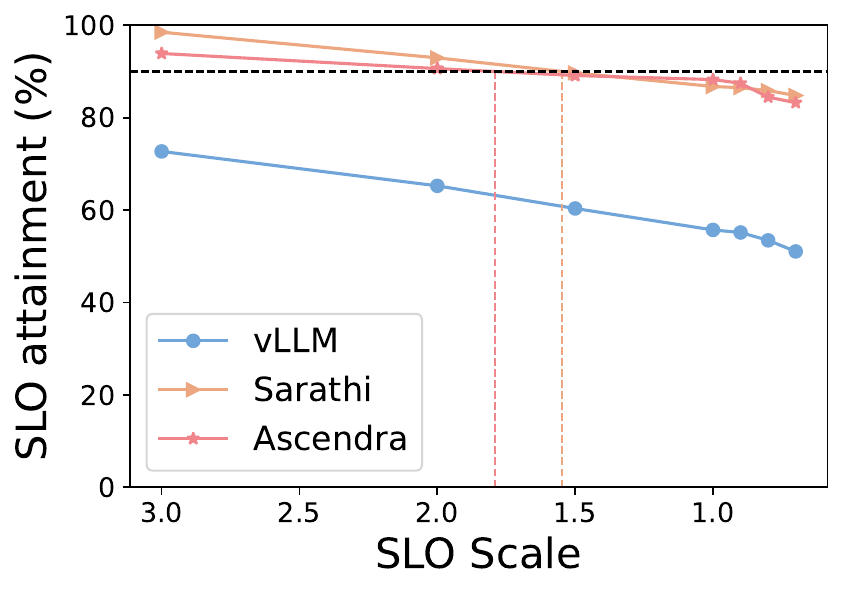}
        
        {\small Qwen-14B (\textit{Openchat-ShareGPT4})}
        \label{fig:qwen_gpt_scale}
    \end{minipage}

    \caption{\textit{Top row}: SLO attainment for \name, vLLM, and Sarathi across three LLM models and datasets. \name consistently increases system capacity to serve requests within SLOs, while vLLM lacks SLO awareness and Sarathi sacrifices TTFT for TBT. \textit{Bottom row}: SLO attainment as the SLO target varies. \name significantly improves goodput and outperforms baselines across a range of SLO thresholds.}
    \label{fig:goodput_main_paper}
\end{figure*}

In LLM serving systems, both TTFT and TBT are critical performance metrics. Following prior work~\cite{zhong2024distserve}, we define goodput as the percentage of requests that meet their TTFT SLO and have a mean TBT that satisfies the TBT SLO. Our objective is to ensure that the system maintains at least 90\% goodput across varying system loads. \autoref{fig:goodput_main_paper} shows a comparative evaluation of goodput for \name and two baselines, vLLM and Sarathi-Serve, under increasing query rates. We also analyze how these systems perform under different SLO conditions by fixing the query rate and varying the SLO strictness, demonstrating performance under both stringent and relaxed SLO settings.

In the first row, we observe that \name improves goodput by 19.1\% and 17.4\% over vLLM and Sarathi-Serve respectively when serving Mistral-7B on~\textit{ShareGPT}. For LLaMA3.1-8B on \textit{LongBench}, \name achieves a 15.4\% improvement compared to both baselines.

\begin{figure*}[t]
    \centering

    \begin{subfigure}{0.495\textwidth}
        \centering
        \includegraphics[width=\linewidth]{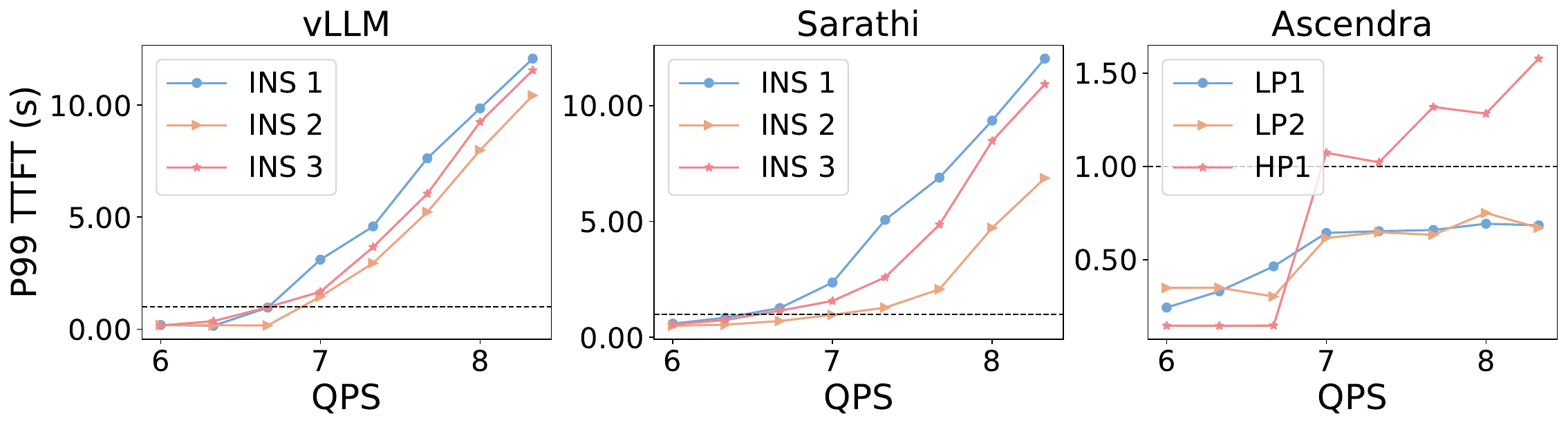}
        \caption{P99 TTFT}
        \label{fig:p99_ttft}
    \end{subfigure}
    \hfill
    \begin{subfigure}{0.495\textwidth}
        \centering
        \includegraphics[width=\linewidth]{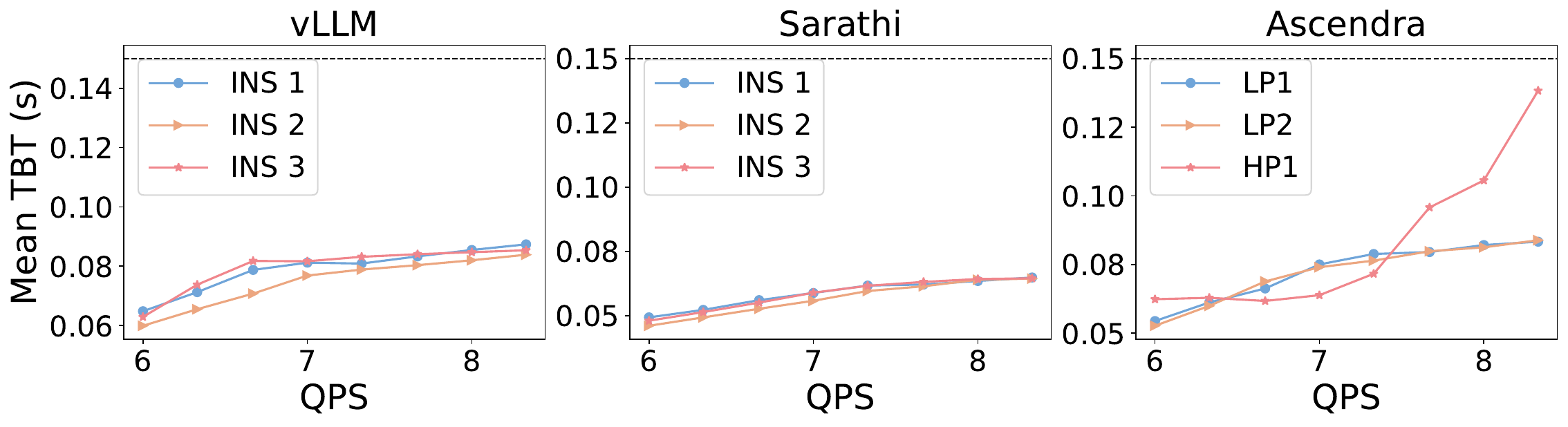}
        \caption{Mean TBT}
        \label{fig:mean_tbt}
    \end{subfigure}

    \begin{subfigure}{0.495\textwidth}
        \centering
        \includegraphics[width=\linewidth]{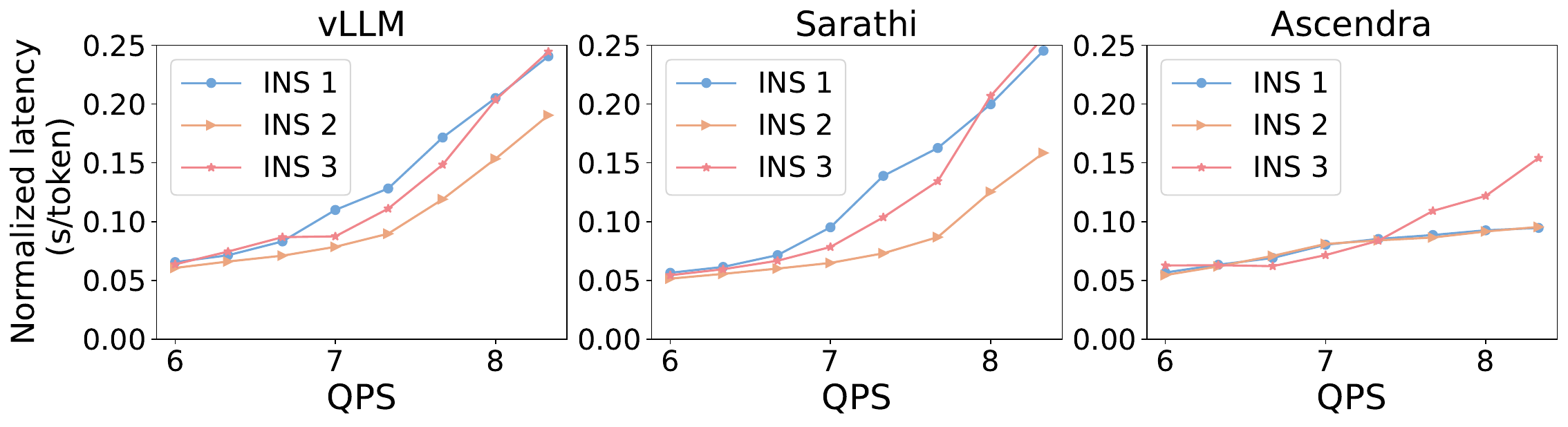}
        \caption{Throughput}
        \label{fig:throughput}
    \end{subfigure}
    \hfill
    \begin{subfigure}{0.495\textwidth}
        \centering
        \includegraphics[width=\linewidth]{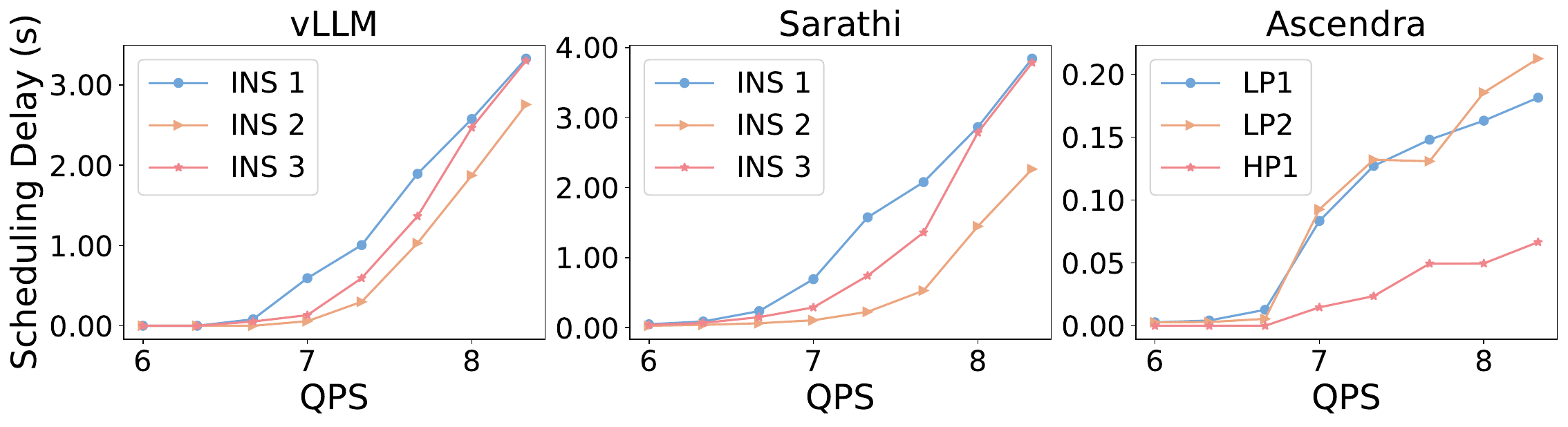}
        \caption{Request scheduling delay}
        \label{fig:scheduling_delay}
    \end{subfigure}
    \vspace{-3mm}

   \caption{Performance breakdown of \name and baselines when serving Mistral-7B on the \textit{ShareGPT} dataset. All systems are deployed with three instances: vLLM and Sarathi-Serve use identical configurations for all instances (INS 1, 2, 3), while \name employs a heterogeneous setup with two LP instances and one HP instance. We report four key metrics: (a) P99 TTFT, (b) Mean TBT, (c) System throughput, and (d) Request scheduling delay of all instances.}
    \label{fig:lp_hp_compare}
\end{figure*}

The performance improvements can be attributed to several key factors. Under high load, the FCFS-based strategies used by vLLM and Sarathi-Serve introduce significant scheduling delays (\autoref{fig:scheduling_delay}), causing many requests to miss their TTFT SLOs. In contrast, \name dynamically reorders incoming requests to prioritize those with tighter TTFT constraints. Requests that have been waiting and are close to violating their TTFT SLO are offloaded to the HP instance, where \name gives them priority, minimizing the additional delay caused by backlogs at the LP instance. However, as the request rate continues to rise, the queue at the HP instance begins to build up, eventually causing some requests to miss their TTFT SLOs and resulting in a drop in goodput. Despite this, \name continues to outperform both baselines even under extreme load, as the LP instance consistently ensures a number of requests meet their SLOs.

In contrast to Mistral-7B and LLaMA3.1-8B, we observe slightly different behavior with the Qwen-14B model. Although vLLM supports a maximum throughput of only 2.55 QPS, both Sarathi-Serve and \name maintain 2.8 QPS. The reason for this lies in the limited available GPU memory when serving Qwen-14B. This model has twice as many parameters as Mistral-7B, leaving significantly less memory available for the KV-cache. Additionally, the per-token KV-cache size increases from 512 KB to 800 KB, further exacerbating memory pressure. As a result, the number of available KV-cache blocks drops by 88\% (from 25,000 to 3,000).

With such limited memory, the optimal scheduling strategy shifts toward minimizing prefills and prioritizing decoding steps to free memory for incoming requests. vLLM performs poorly in this setting because it eagerly executes prefills, quickly exhausting memory and either preempting ongoing decodes or delaying pending prefills. Sarathi-Serve performs better by piggybacking decodes with prefills, allowing some decodes to complete and release memory, thus making space for additional prefills. Similarly, \name performs on par with Sarathi-Serve, as the LP instance in \name naturally prioritizes finishing decode requests for higher throughput. This experiment demonstrates that \name remains competitive even in extreme scenarios with extremely limited memory.


The second row of \autoref{fig:goodput_main_paper} demonstrates the flexibility of \name under varying SLO constraints. In this experiment, we fix the request arrival rate at the point where \name achieves slightly above 90\% goodput. Then, we vary the TTFT and TBT thresholds from \autoref{tab:model_comparison} to apply more stringent and more relaxed SLOs. Across all models and datasets, \name consistently maintains higher goodput than the baselines. Under a certain SLO guarantee (90\%), \name can sustain up to 4x more stringent SLO scale.

\subsection{Effectiveness of LP and HP Instances}
Each instance type in \name is optimized for a specific goal: LP instances prioritize low TTFT and high throughput, while HP instances minimize scheduling delay for urgent requests. We evaluate how well each instance achieves its design objectives using the setup from \autoref{tab:model_comparison}, measuring TTFT, TBT, throughput, and scheduling delay (\autoref{fig:lp_hp_compare}).

As shown in \autoref{fig:p99_ttft}, \name significantly reduces TTFT violations compared to baselines, demonstrating the effectiveness of LP instances in proactively offloading urgent requests to HP instances, even under high load. This allows \name to maintain stable goodput rather than wasting compute on requests that have already missed their SLOs. \autoref{fig:mean_tbt} further shows that both LP and HP instances consistently keep TBT within SLO, contributing to overall goodput gains.

Since HP instances focus on minimizing latency, they naturally achieve lower scheduling delay, as seen in \autoref{fig:scheduling_delay}, where HP requests wait $4\times$ less than LP requests. However, this comes at the cost of throughput, as HP follows strict FCFS without the flexibility to cherry-pick requests like LP. These results highlight the key trade-off: HP instances enable timely execution of urgent requests at the expense of slightly reduced throughput.

\subsection{Choice of Reorder Policy} \label{subsec:order_policy}
\begin{figure}[t]
    \centering
    \includegraphics[width=.495\columnwidth]
    {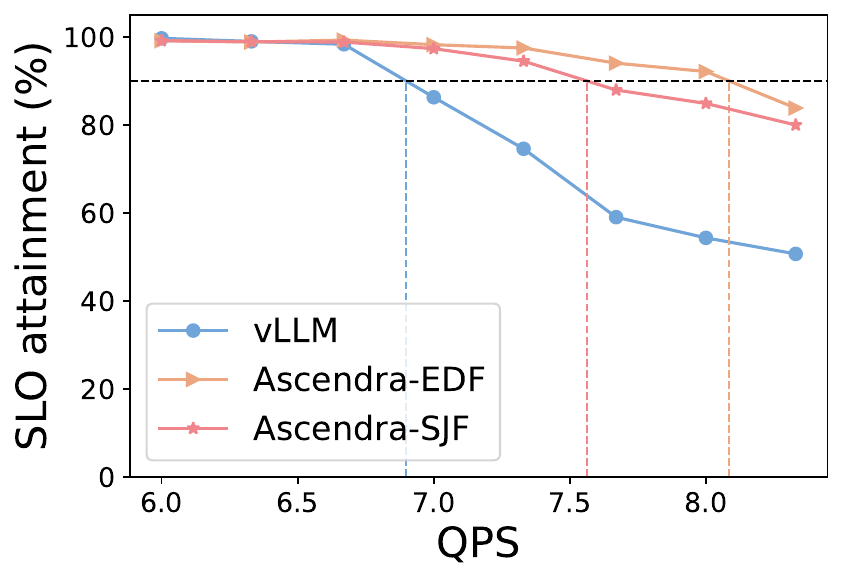}
    \includegraphics[width=.495\columnwidth]
    {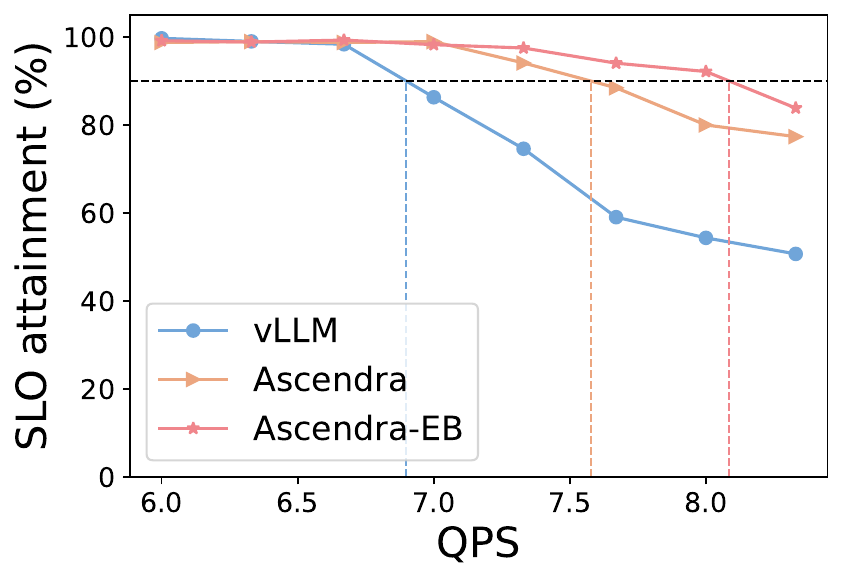}
    \caption{\textit{Left}: Different value function policies result in different out of order scheduling schemes, leading to different goodputs. \textit{Right}:Elastic batch size on HP instance further increase the goodput under high QPS.}
    \label{fig:ablation_eb_val}
\end{figure}

To support diverse application objectives and service-level differentiation, we introduce a configurable reordering policy mechanism within \name, as described in \autoref{subsec:out_of_order}. This feature allows operators to tailor the execution order of queued requests on LP instances based on application-specific priorities or workload characteristics. In the left panel of \autoref{fig:ablation_eb_val}, we evaluate two policy variants: SJF, which prioritizes requests with shorter expected execution times, and EDF, the default policy in \name, which favors requests closer to their TTFT deadline.

Our results show that \name-EDF achieves higher goodput than \name-SJF in this scenario. This highlights the importance of aligning the scheduling policy with the workload characteristics and target SLOs. More broadly, it demonstrates the flexibility of \name to adapt to different operational goals. For instance, operators can configure policies to favor low-latency responses for latency-sensitive applications (i.e., stringent TTFT SLO), or to prioritize premium users over free-tier users in multi-tenant environments. This policy-driven adaptability enables \name to serve as a versatile foundation for building fair and efficient LLM serving systems across diverse deployment settings.

\subsection{Effectiveness of Elastic Batch}
We demonstrate the effectiveness of elastic batch size. For comparison, we use the same setup: 2 LP instances and 1 HP instance, with the Mistral-7B model on the \textit{ShareGPT} dataset. We compare the system goodput with and without enabling elastic batch size on the HP instance. \autoref{fig:ablation_eb_val} (right panel) shows that dynamic scaling of the batch size further improves system goodput. This improvement happens because when the amount of free memory exceeds 10\% of the total memory, the system can safely increase the batch size. A larger batch size allows the HP instance to admit and process more prefill requests concurrently, which in turn leads to a higher proportion of requests meeting their TTFT SLOs and improves overall throughput.

\begin{figure}[t]
    \centering
    \includegraphics[width=\columnwidth]
    {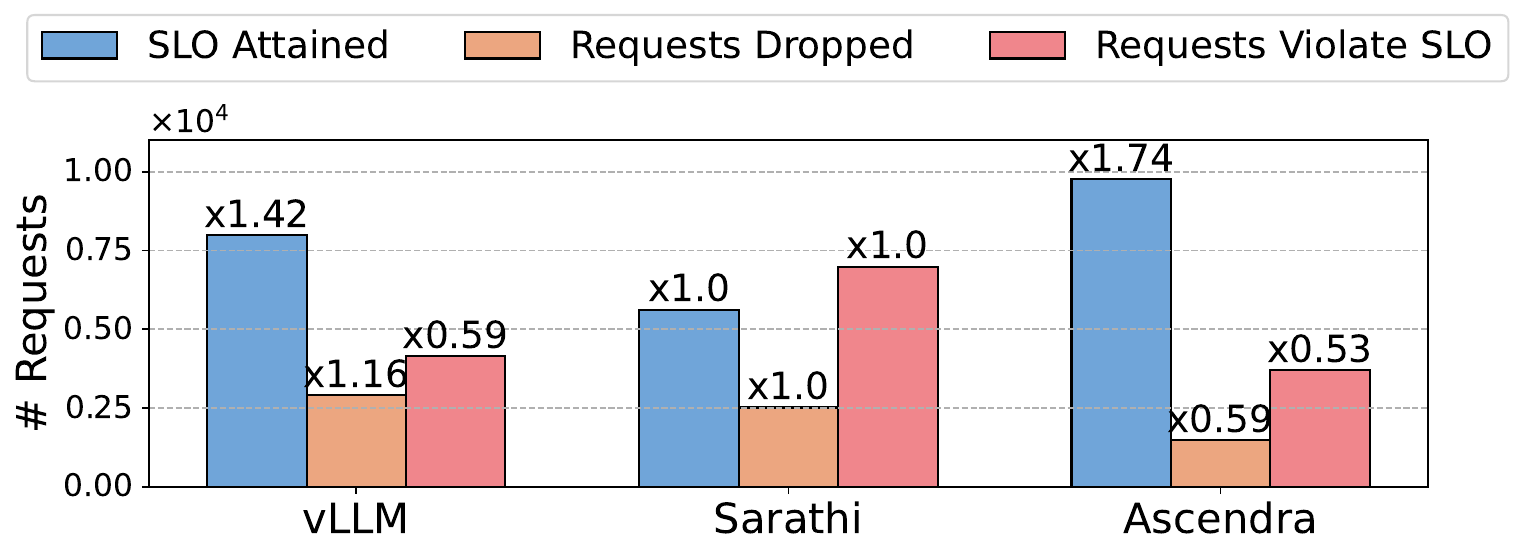}
    \caption{\name serves the most request under SLO with only a small fraction of request being dropped.}
    \label{fig:drop_llama}
    \vspace{-3 mm}
\end{figure}

\subsection{Serving under Heavy System Load}
In real-world deployment, system load can fluctuate and occasionally exceed capacity. A common strategy to prevent backlog build-up is to drop requests that violate SLOs~\cite{shepherd}. We evaluated all frameworks by running them for 10 minutes at an arrival rate slightly above system capacity. Any request that remains in the waiting queue past the TTFT SLO is dropped. \autoref{fig:drop_llama} shows the comparison across all baselines. Under the same load and time period, \name serves the 1.74x more request under SLO comparing to baseline methods while reduce the number of requests dropped and requests violates SLO by almost 50\%.

\subsection{Scaling \name to Cluster of GPUs}
In this section, we discuss how \name can be scaled to large GPU clusters. When deploying across hundreds or thousands of GPUs, searching for the optimal system configuration can be both time-consuming and resource-intensive. To reduce the configuration search space and enable scalable deployment on the fly, we scale the system using subgroups. 

\begin{figure}[t]
    \centering
    \includegraphics[width=.495\columnwidth]
    {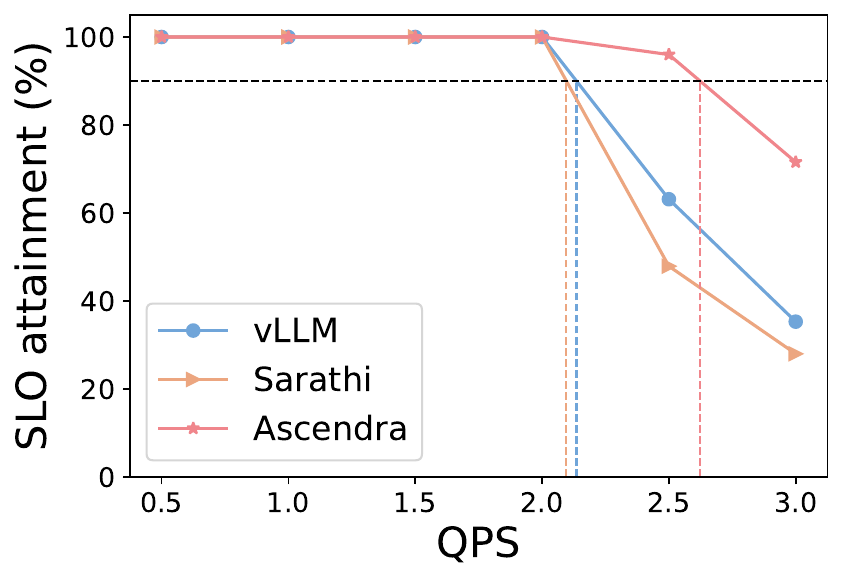}
    \includegraphics[width=.495\columnwidth]
    {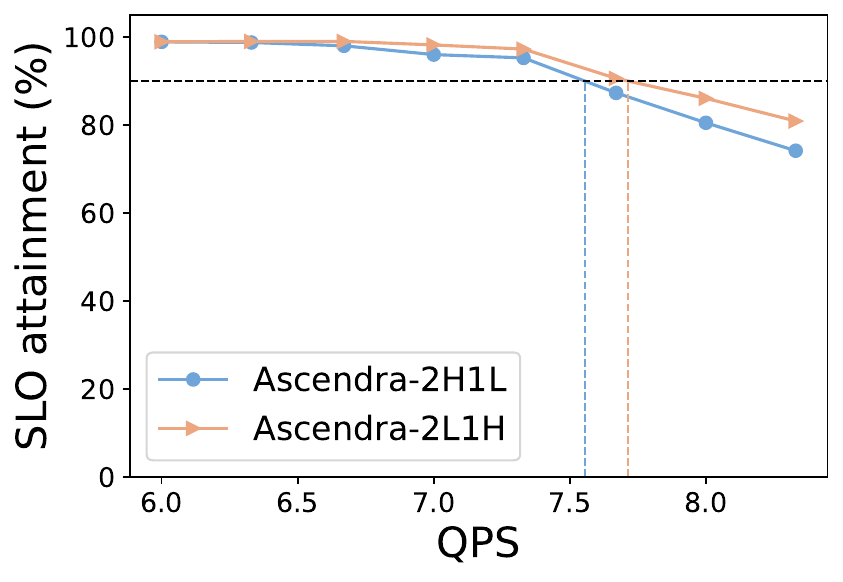}
    \caption{\text{Left}: Goodput comparison of \name and baselines on \textit{LongBench} dataset with two instances setup, one LP instance and one HP instance. \textit{Right}: Goodput comparison for two different configurations of three instances setup: two LP instances and one HP instance vs one LP instance and two HP instances on \textit{ShareGPT4} dataset.}
    \label{fig:scale_fig}
    \vspace{-3mm}
\end{figure}

In \autoref{fig:goodput_main_paper}, we show that for a subgroup of three instances with two LP and one HP instance (2L1H), \name outperforms existing baselines. \autoref{fig:scale_fig} further compares the performance of different subgroup configurations, including a 2H1L setup (two HP and one LP instance) and a two-instance subgroup with one LP and one HP instance (1L1H). We see that for subgroups of three instances, the configuration of 2L1H performs better than 2H1L, Additionally, the 1L1H subgroup also achieves better performance than the baselines, highlighting the general effectiveness. These results suggest that when deploying \name on a large GPU cluster, the system can be partitioned into independent subgroups of two or three instances for scalable and efficient operation. While these configurations outperform existing baselines, they do not necessarily represent the global optimum. Given sufficient time and knowledge of workload characteristics, a global optimal configuration could be discovered via combinatorial search through our simulator.
\section{Related Work}
Efficiently serving LLMs presents several challenges, including managing long prompts, meeting strict latency requirements (e.g., TTFT and TBT), and maximizing GPU utilization under memory constraints. A wide range of systems have been proposed to address these challenges at different layers of the stack.
At the kernel level, systems such as FlashAttention~\cite{dao2022flashattention} and Flash-Decoding~\cite{hong2024flashdecoding++} optimize core GPU operators to speed up inference. These approaches are orthogonal to ours and can be integrated into most serving pipelines.

At the infrastructure level, systems like AlpaServe~\cite{li2023alpaserve} exploit parallelism in deep neural networks to improve throughput, while SpotServe~\cite{miao2024spotserve} leverages preemptible instances to reduce monetary cost. Orca~\cite{orca} introduces \textit{continuous batching}, allowing the serving engine to interleave prefill and decoding phases across requests. vLLM~\cite{kwon2023efficient} introduces a systematic memory management scheme via \textit{PagedAttention}. Similarly, NanoFlow~\cite{zhu2024nanoflow} overlaps communication and computation to further improve throughput. ExeGPT~\cite{exegpt} leveraging distribution of input and output sequence to dynamically determine the optimal execution configuration for high throughout. While these systems focus on maximizing GPU utilization and system throughput, they generally do not account for application-level TTFT and TBT SLO.

To better meet latency SLOs, systems such as Sarathi-Serve~\cite{agrawal2024taming} and DeepSpeed-FastGen~\cite{holmes2024deepspeed} introduce \textit{chunked prefill}, splitting the prefill phase into smaller segments and \textit{piggybacking decoding} to improve TBT at the expense of higher TTFT. To address both TTFT and TBT, DistServe~\cite{zhong2024distserve}, SplitWise~\cite{patel2024splitwise}, and Llumnix~\cite{sun2024llumnix} support request migration across instances to mitigate prefill-decode interference, enable horizontal scaling, and alleviate memory fragmentation. However, these approaches incur significant communication overhead due to large KV-cache transfers.

Separately, a line of work has focused on KV-cache optimization. CachedAttention~\cite{gao2024cost} proposes a hierarchical storage scheme that places KV tensors in CPU memory or disk. LoongServe~\cite{wu2024loongserve} builds a distributed KV-cache pool across GPUs to support longer prompts. ALISA~\cite{zhao2024alisa} explores the trade-off between recomputing and storing KV tensors in CPU memory, while CacheGen~\cite{liu2024cachegen} compresses KV tensors to reduce memory footprint at the expense of accuracy. While these methods reduce KV-cache size, our approach aims to avoid KV-cache transfer altogether, mitigating inter-instance communication overhead. These techniques are complementary to ours and can be integrated to further improve system performance.

\section{Conclusion}\label{sec:conclusion}
In this paper, we introduced \name, a novel LLM serving system designed to maximize goodput across diverse workloads through a two-tier instance architecture: Low-Priority (LP) and High-Priority (HP) instances. LP instances employ out-of-order scheduling guided by configurable value functions to form efficient batches, while proactively offloading requests at risk of violating their deadlines to HP instances. HP instances, in turn, offer low-latency execution for urgent requests, trading off some throughput to ensure SLO compliance. This collaborative design enables \name to dynamically balance latency and throughput, thus significantly improving overall system goodput. Extensive evaluations using multiple open-source models and representative datasets show that \name consistently outperforms state-of-the-art LLM serving systems.
\newpage

\bibliographystyle{ACM-Reference-Format}
\bibliography{reference}

\clearpage

\appendix

\section{Appendix} \label{sec:appendix}
\subsection{Model Architecture and Batch Request Symbols Definition}
First, we define the symbols related to the architecture of the LLM model below:
\begin{itemize}
    \item $\mathbf{h}$: hidden size
    \item $\mathbf{n}$: number of heads
    \item $\mathbf{s}$: head size ($\mathbf{h} = \mathbf{n} \cdot \mathbf{s}$)
    \item $\mathbf{m}$: FFN intermediate size
\end{itemize}
\textbf{Note:} If tensor parallelism is used, $h$, $n$, and $m$ should be divided by the tensor parallelism size.

Second, we define the symbols that characterize the batch to be executed:
\begin{itemize}
     \item $B_{p}, B_{d}$: number of prefill requests in the batch, number of decode requests in the batch, $B =B_{p}+B_{d}$.
    \item $p_0, p_1, \dots, p_{B_p-1}$: input prompt length for each prefill request in the batch.
    \item $l_0, l_1, \dots, l_{B_{p}-1}$: number of prompt tokens processed up till the current batch of each prefill request in the batch.
    \item $c_0, c_1, \dots, c_{B_{p}-1}$: number of prefill tokens (chunk size) to be processed in the current batch.
    \item $\hat{l}_0, \hat{l}_1, \dots, \hat{l}_{B_{d}-1}$: input prompt length plus the number of decoded token generated up to the current batch of each decode request.
    \item $t$: number of prompt tokens of the prefill requests in the batch, $t = \sum\limits_{i=0}^{B_p-1} p_i$
    \item $b$: block size in the attention kernel. This parameter is used in FlashAttention~\cite{dao2022flashattention}, a common kernel optimization technique adopted by current LLM serving systems (VLLM, Sarathi-Serve).
\end{itemize}

\subsection{Prefill Analysis (w.o chunking)}~\label{subsec:prefill_analysis}
We first consider the general matrix multiply(GEMM) operations. Considering a request with a prompt token length of $p$, the GEMM-related computational flops and memory access are presented in Table~\ref{tbl:prefill_gemm}. We refer to fetching data from HBM to SRAM as memory access.

\begin{table*}[htbp]  
    \centering
    \begin{tabular}{|c|c|c|c|c|c|}
    \hline
    \textbf{GEMM Name} & \textbf{Shape of M} & \textbf{Shape of N} & \textbf{FLOPs} & \textbf{Memory Read} &  \textbf{Memory Write}\\
    \hline
    QKV Linear & $(t, h)$ & $(h, 3h)$ & $3th^2$ & $th+3h^2$ & $3th$ \\
    \hline
    Attn Output & $(t, h)$ & $(h, h)$ & $th^2$ & $th+h^2$ & $th$\\
    \hline
    FFN Input & $(t, h)$ & $(h, m)$ & $thm$ & $th+hm$ & $tm$\\
    \hline
    FFN Output & $(t, m)$ & $(m, h)$ & $thm$ & $tm+hm$ & $th$ \\
    \hline
    \end{tabular}
    \caption{GEMM operations' FLOPs and memory access in a prefill batch.}
    \label{tbl:prefill_gemm}
\end{table*}

\begin{table*}[htbp]
\centering
\begin{tabular}{|c|c|c|c|c|c|}
\hline
\textbf{GEMM Name} & \textbf{Shape of M} & \textbf{Shape of N} & \textbf{FLOPs} & \textbf{Memory Read} &  \textbf{Memory Write}\\
\hline
QKV Linear & $(B_{d}, h)$ & $(h, 3h)$ & $3B_{d}h^2$ & $B_{d}h+3h^2$ & $3B_{d}h$ \\
\hline
Attn Output & $(B_{d}, h)$ & $(h, h)$ & $B_{d}h^2$ & $B_{d}h+h^2$ & $B_{d}h$\\
\hline
FFN Input & $(B_{d}, h)$ & $(h, m)$ & $B_{d}hm$ & $B_{d}h+hm$ & $B_{d}m$\\
\hline
FFN Output & $(B_{d}, m)$ & $(m, h)$ & $B_{d}hm$ & $B_{d}m+hm$ & $B_{d}h$ \\
\hline
\end{tabular}
\caption{GEMM operations' FLOPs and memory access in a decode batch.}
\label{tbl:decode_gemm}
\end{table*}

For the attention operation, we discuss the attention operation with FlashAttention~\cite{dao2022flashattention} optimization. Given a request with $p$ prompt tokens, inputs $\mathbf{Q}$, $\mathbf{K}$, $\mathbf{V} \in \mathbb{R}^{p \times s}$, the attention computation is expressed follows:
\[
\mathbf{S} = \mathbf{Q} \mathbf{K}^\top \in \mathbb{R}^{p \times p}, 
\mathbf{P} = \text{softmax}(\mathbf{S}) \in \mathbb{R}^{p \times p},
\mathbf{O} = \mathbf{P} \mathbf{V} \in \mathbb{R}^{p \times s}
\]
We refer to the forward pass algorithm 2 in FlashAttention~\cite{dao2022flashattention} for the following IO operation analysis. First, every K, V block is loaded only once from HBM to SRAM, which counts for $2 s p$ memory read access. For each K, V block pair, the corresponding Q, O blocks need to be loaded, performing the attention computation, and write back to O. This accounts for the addition of $2 s p$ memory read access and $s p$ memory write access. Rather than operating on each individual K,V pair, FlashAttention operated on blocks of K, V pairs with block size $b$. The total number of K, V block pairs can be expressed as $p/b$. Therefore, the total memory access for prefill with prompt token length $p$ is $2 s p + 3 s p \cdot (p/b)$. For flops computations, the multiplication of the matrix of $QK^{T}$ and $PV$ yields $2p^2 s$ flops.
Therefore, together with the GEMM computation and memory access, the computation and memory access of a batch with $B_{p}$ prefill requests can be represented as:
\begin{equation}
    \begin{aligned}
        M_{p} &= 4h^2+8\sum\limits_{i=0}^{B_{p}-1}p_{i}h+2hm+2tm + 2sp+3sp\cdot\frac{p}{b} \\
        F_{p} &= 4\sum\limits_{i=0}^{B_{p}-1}p_{i}h^2 + 2\sum\limits_{i=0}^{B_{p}-1}p_{i}hm + 2p^2s
    \end{aligned}
\end{equation}

\subsection{Decode Analysis}\label{subsec:decode_analysis}
Similarly, for decode analysis, we also first consider GEMM operations. The computational flops and the memory access of the GEMM operations are presented in Table~\ref{tbl:decode_gemm}.
For the attention operation of decode, assuming a request with 1 prefill token only, to decode the first token, the related $k,v \in R^{s}$ of the first token needs to be fetched along with the query vector $q\in R^{s}$; for the second generated token, we need to fetch the $k,v \in R^{s}$ of the first generated token as well as the second query vector $q\in R^{s}$, etc. Therefore, to generate $p$ tokens, it requires $3sp$ memory access. However, this only holds true if there is only one decoding request in a batch and the total number of prefill token and decode token does not exceed the SRAM capacity.

 To generalize this, we consider the batched scenario in which multiple decode requests exist. Their memory may exceed the SRAM capacity. Thus, to decode a single token from a request, one may need to reload $k,v  \in R^{p\times s}$ of the prefilled prompt as well as the previously generated tokens into SRAM. In such a scenario, the memory access during decode attention depends on the number of requests in the current batch as well as their individual related $K,V$ size. However, we can derive an upper bound of this memory access. 
 
 Assuming that for every decoding token of a request, we need to reload the $k,v$ of its prefilled prompt as well as all tokens generated up to now. For a request $i$ with $\hat{l}_{i}$ prefill tokens and generated tokens, the memory access of all decoding request in this batch is bounded by 
$
\sum\limits_{i=0}^{i=B_{d}-1}2\hat{l}_{i}s+2s
$,
where 2s represents the $k,v\in R^{s}$ for the generated token.
For the flops computation of the decode batch, the new query vector $q \in R^{s}$ needs to compute its attention against all prefill and generated $k,v$, which can be expressed as
$
\sum\limits_{i=0}^{i=B_{d}-1}2\hat{l}_{i}s
$.

Therefore, together with the GEMM computation and memory access, the computation and memory access of a batch with $B_{d}$ decode requests can be represented as:
\begin{equation}
    \begin{aligned}
M_{d}= 4h^2+8Bh+2hm+2Bm+\sum\limits_{i=0}^{i=B_{d}-1}2\hat{l}_{i}s+2s\\
F_{d}= 4Bh^2+2Bhm+\sum\limits_{i=0}^{i=B_{d}-1}2\hat{l}_{i}s
    \end{aligned}
\end{equation}

\subsection{Hybrid Batch with Chunked Prefill}~\label{app:analytical_modeling}
In a hybrid batch with chunked prefill, the GEMM operations if the prefill request and decode requests can be derived from table~\ref{tbl:prefill_gemm}\&~\ref{tbl:decode_gemm} individually. Similarly, the memory access and flops calculation for the deoding request still is bounded by the result derived in~\ref{subsec:decode_analysis}. However, the memory access for the chunked prefill requests is closely related to the number of processed prompt tokens of each request up to the current batch. 

Suppose that there are $B_{p}$ prefills in a hybrid batch. The total number of prefill tokens processed in the batch can be expressed as
$
\sum\limits_{i=0}^{B_{p}-1} c_{i}
$.
To compute the attention for prompt chunk of each request in the current batch, we need to load all previously processed $k,v \in R^{l \times s}$ of each request in the current batch. This accounts for $2 l_{i} s$ memory access for each request. Similarly to~\ref{subsec:prefill_analysis}, for every $k,v$ block pair, we need to load the $q$ block to compute the attention. However, the difference from~\ref{subsec:prefill_analysis} is that we only need to access the $q$ block of the current chunk. The total memory access for each $k,v$ block pair is $3 c_{i} s$, which accounts for the memory read, write for the $q,o$ block. The total number of $k,v$ block pairs is equal to $l_{i}/b$. So the memory access of the prefill tokens in the current hybrid batch is expressed as
$
\sum\limits_{i=0}^{B_{p-1}}2l_{i}s+3c_{i}s(l_{i}/b)
$. Regarding the count of flops, the query $q\in R^{c_{i} \times s}$ in the current batch of request $i$ needs to perform attention against all previous processed $K,V$ tokens of $l_{i}$. The amount of flops in $P=qK^{T}$ can be expressed as $sl_{i}c_{i}, q\in R^{c_{i}\times s}, K\in R^{l_{i}\times s}$, and in $O=PV$ can be expressed as $sl_{i}c_{i}, P\in R^{c_{i} \times l_{i}}, V\in R^{l_{i}\times s}$. So, to prefill all query tokens for request $i$ in the current batch, the total flops can be expressed as $2sl_{i}c_{i}$. So in a hybrid batch with $B_{p}$ chunked prefill requests, the flops count can be expressed as 
$
\sum\limits_{i=0}^{B_{p}-1}2sl_{i}c_{i}
$.
Together with the decode request $B_{d}$ in the batch, we express the total memory access in the hybrid batch as 
\begin{align}
M_{m} &= 4h^2+2hm 
+ \Bigg( 8\sum\limits_{i=0}^{B_{p-1}}l_{i}h 
+ 2\sum\limits_{i=0}^{B_{p-1}}l_{i}m 
+ \sum\limits_{i=0}^{B_{p-1}}2l_{i}s  \notag \\
&\quad + 3c_{i}s\frac{l_{i}}{b} \Bigg) 
+ \Bigg( 8B_{d}h + 2B_{d}m 
+ \sum\limits_{i=0}^{B_{d}-1}2\hat{l}_{i}s + 2s \Bigg).
\end{align}

The corresponding flops can be expressed as 
\begin{align}
F_{m} &= \Bigg( \sum\limits_{i=0}^{B_{p}-1} 
\big( 4c_{i}h^{2} + 2c_{i}hm + 2sl_{i}c_{i} \big) \Bigg)  \notag \\
&\quad + \Bigg( 4B_{d}h^2 + 2B_{d}hm 
+ \sum\limits_{i=0}^{B_{d}-1} 2\hat{l}_{i}s \Bigg).
\end{align}

\subsection{Goodput Comparison}\label{app: appendix_result}
\begin{figure*}[t]
    \centering
    \begin{minipage}{0.32\textwidth}
        \centering
        \includegraphics[width=\linewidth]{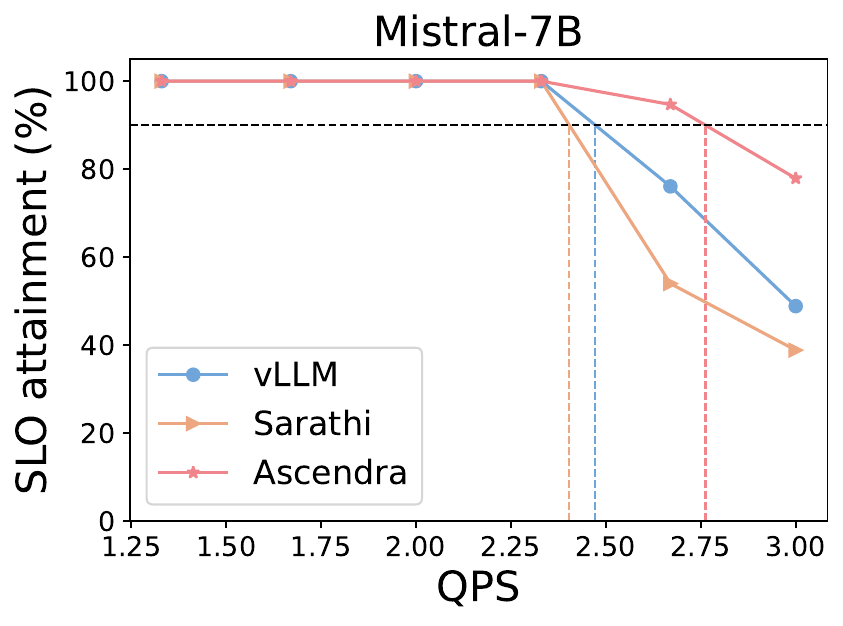}
        \label{fig:mistral_longbench}
    \end{minipage}
    \hfill
    \begin{minipage}{0.32\textwidth}
        \centering
        \includegraphics[width=\linewidth]{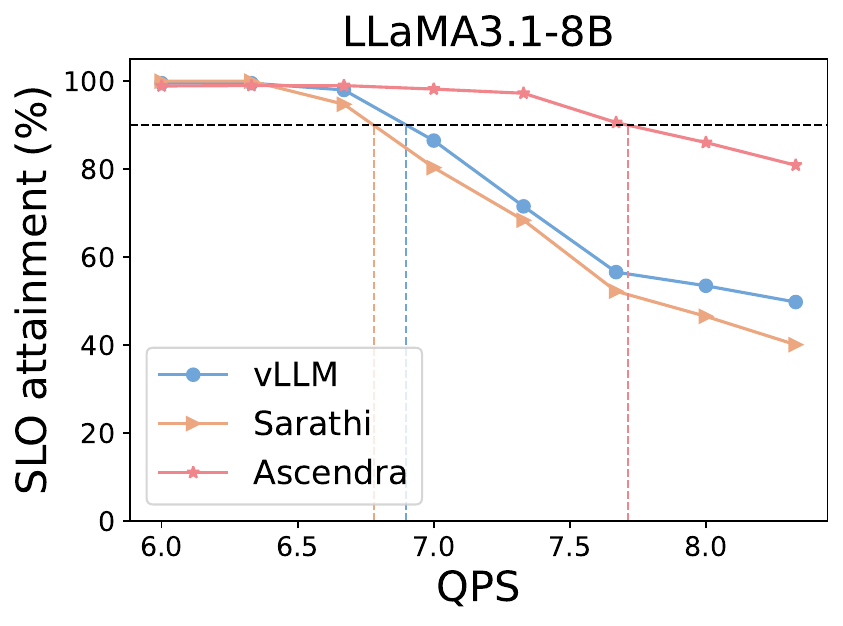}
        \label{fig:llama_gpt}
    \end{minipage}
    \hfill
    \begin{minipage}{0.32\textwidth}
        \centering
        \includegraphics[width=\linewidth]{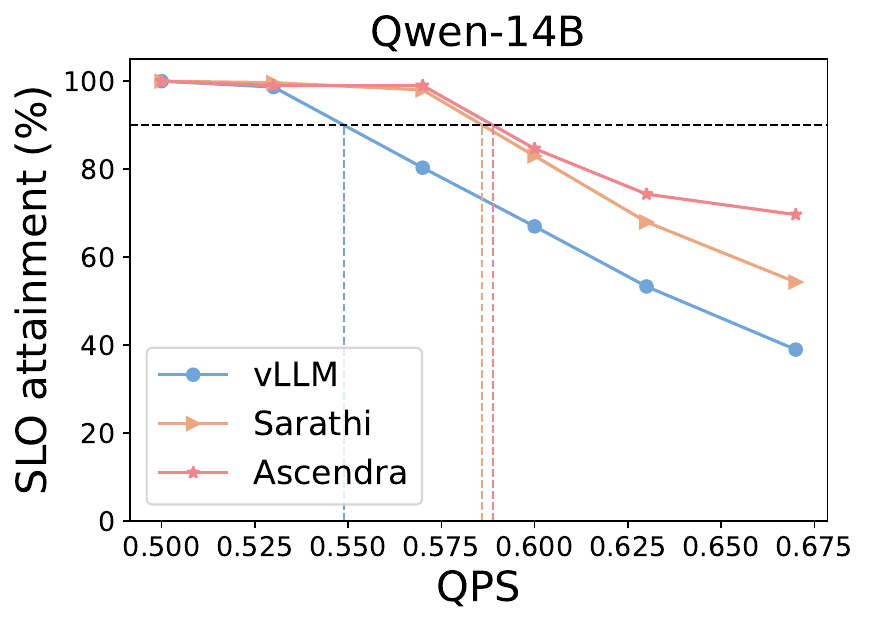}
        \label{fig:qwen_longbench}
    \end{minipage}

    \begin{minipage}{0.32\textwidth}
        \centering
        \includegraphics[width=\linewidth]{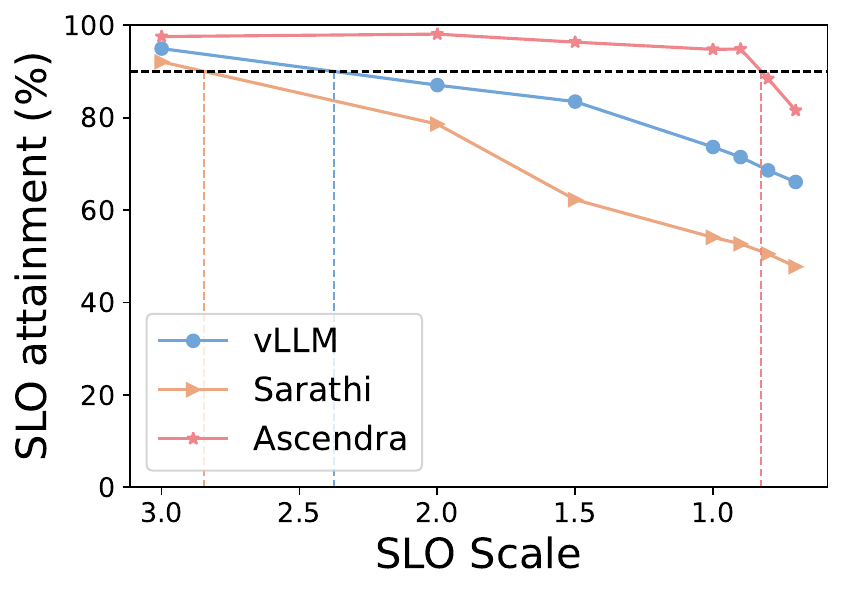}
        
        {\small Mistral-7B (\textit{LongBench})}
        \label{fig:mistral_longbench_scale}
    \end{minipage}
    \hfill
    \begin{minipage}{0.32\textwidth}
        \centering
        \includegraphics[width=\linewidth]{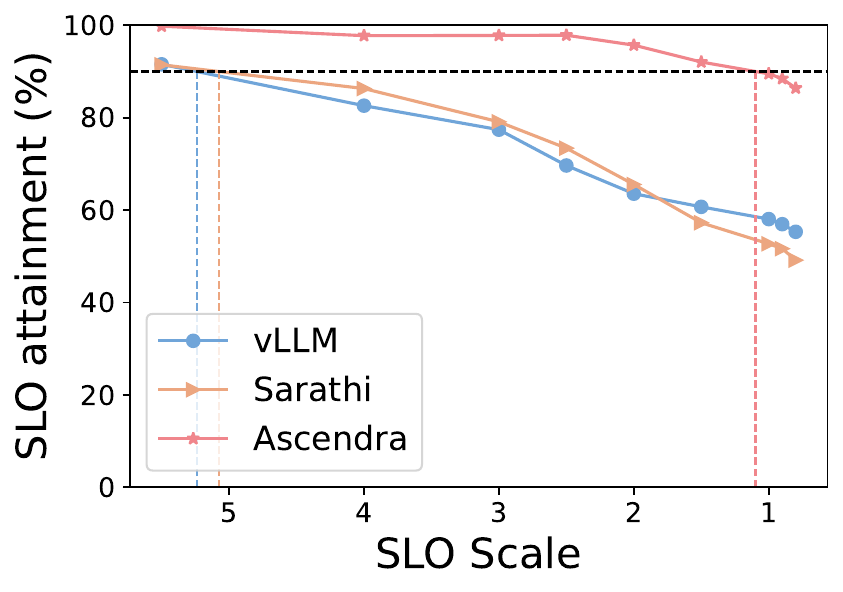}
        
        {\small LLaMA3.1-8B (\textit{Openchat ShareGPT4})}
        \label{fig:llama_gpt_scale}
    \end{minipage}
    \hfill
    \begin{minipage}{0.32\textwidth}
        \centering
        \includegraphics[width=\linewidth]{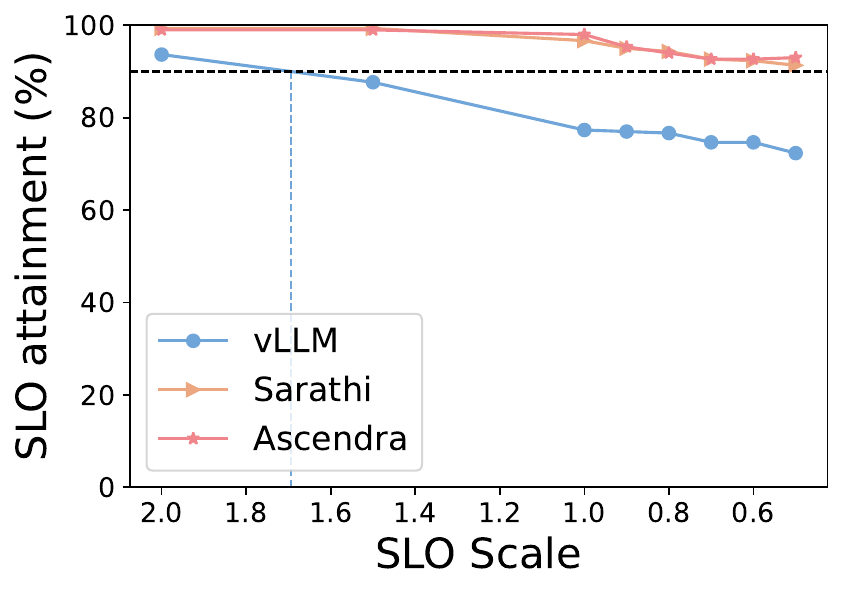}
        
        {\small Qwen-14B (\textit{LongBench})}
        \label{fig:qwen_longbench_scale}
    \end{minipage}

    \caption{Top row shows the goodput comparison for \name, vLLM, Sarathi-Serve with Mistral-7B, LlaMA3.1-8B and Qwen-14B on \textit{Openchat ShareGPT4} and \textit{Longbench} text summarization dataset. Bottom row shows the flexibility of \name with respect to the scale of the SLOs.}
    \label{fig:goodput_main_appendix}
\end{figure*}

\autoref{fig:goodput_main_appendix} shows the goodput performance of models on different datasets that is not presented in the main paper.

\end{document}